\PassOptionsToPackage{unicode}{hyperref}
\PassOptionsToPackage{hyphens}{url}
\PassOptionsToPackage{dvipsnames,svgnames,x11names}{xcolor}
\documentclass[
  11pt,
  letterpaper,
  DIV=11,
  numbers=noendperiod]{scrartcl}
\usepackage{xcolor}
\usepackage[margin=1in]{geometry}
\usepackage{amsmath,amssymb}
\usepackage{iftex}
\ifPDFTeX
  \usepackage[T1]{fontenc}
  \usepackage[utf8]{inputenc}
  \usepackage{textcomp} 
\else 
  \usepackage{unicode-math} 
  \defaultfontfeatures{Scale=MatchLowercase}
  \defaultfontfeatures[\rmfamily]{Ligatures=TeX,Scale=1}
\fi
\usepackage{lmodern}
\ifPDFTeX\else
\fi
\IfFileExists{upquote.sty}{\usepackage{upquote}}{}
\IfFileExists{microtype.sty}{
  \usepackage[]{microtype}
  \UseMicrotypeSet[protrusion]{basicmath} 
}{}
\makeatletter
\@ifundefined{KOMAClassName}{
  \IfFileExists{parskip.sty}{%
    \usepackage{parskip}
  }{
    \setlength{\parindent}{0pt}
    \setlength{\parskip}{6pt plus 2pt minus 1pt}}
}{
  \KOMAoptions{parskip=half}}
\makeatother
\makeatletter
\ifx\paragraph\undefined\else
  \let\oldparagraph\paragraph
  \renewcommand{\paragraph}{
    \@ifstar
      \xxxParagraphStar
      \xxxParagraphNoStar
  }
  \newcommand{\xxxParagraphStar}[1]{\oldparagraph*{#1}\mbox{}}
  \newcommand{\xxxParagraphNoStar}[1]{\oldparagraph{#1}\mbox{}}
\fi
\ifx\subparagraph\undefined\else
  \let\oldsubparagraph\subparagraph
  \renewcommand{\subparagraph}{
    \@ifstar
      \xxxSubParagraphStar
      \xxxSubParagraphNoStar
  }
  \newcommand{\xxxSubParagraphStar}[1]{\oldsubparagraph*{#1}\mbox{}}
  \newcommand{\xxxSubParagraphNoStar}[1]{\oldsubparagraph{#1}\mbox{}}
\fi
\makeatother

\usepackage{longtable,booktabs,array}
\usepackage{calc} 
\usepackage{etoolbox}
\makeatletter
\patchcmd\longtable{\par}{\if@noskipsec\mbox{}\fi\par}{}{}
\makeatother
\IfFileExists{footnotehyper.sty}{\usepackage{footnotehyper}}{\usepackage{footnote}}
\makesavenoteenv{longtable}
\usepackage{graphicx}
\makeatletter
\newsavebox\pandoc@box
\newcommand*\pandocbounded[1]{
  \sbox\pandoc@box{#1}%
  \Gscale@div\@tempa{\textheight}{\dimexpr\ht\pandoc@box+\dp\pandoc@box\relax}%
  \Gscale@div\@tempb{\linewidth}{\wd\pandoc@box}%
  \ifdim\@tempb\p@<\@tempa\p@\let\@tempa\@tempb\fi
  \ifdim\@tempa\p@<\p@\scalebox{\@tempa}{\usebox\pandoc@box}%
  \else\usebox{\pandoc@box}%
  \fi%
}
\def\fps@figure{htbp}
\makeatother

\NewDocumentCommand\citeproctext{}{}

\makeatletter
 \let\@cite@ofmt\@firstofone
 \def\@biblabel#1{}
 \def\@cite#1#2{{#1\if@tempswa , #2\fi}}
\makeatother
\newlength{\cslhangindent}
\newlength{\csllabelwidth}
\newenvironment{CSLReferences}[2] 
 {\begin{list}{}{%
  \setlength{\itemindent}{0pt}
  \setlength{\leftmargin}{0pt}
  \setlength{\parsep}{0pt}
  \ifodd #1
   \setlength{\leftmargin}{\cslhangindent}
   \setlength{\itemindent}{-1\cslhangindent}
  \fi
  \setlength{\itemsep}{#2\baselineskip}}}
 {\end{list}}
\usepackage{calc}

\providecommand{\tightlist}{%
  \setlength{\itemsep}{0pt}\setlength{\parskip}{0pt}}

\KOMAoption{captions}{tableheading}
\makeatletter
\@ifpackageloaded{caption}{}{\usepackage{caption}}
\AtBeginDocument{%
\ifdefined\contentsname
  \renewcommand*\contentsname{Table of contents}
\else
  \newcommand\contentsname{Table of contents}
\fi
\ifdefined\listfigurename
  \renewcommand*\listfigurename{List of Figures}
\else
  \newcommand\listfigurename{List of Figures}
\fi
\ifdefined\listtablename
  \renewcommand*\listtablename{List of Tables}
\else
  \newcommand\listtablename{List of Tables}
\fi
\ifdefined\figurename
  \renewcommand*\figurename{Figure}
\else
  \newcommand\figurename{Figure}
\fi
\ifdefined\tablename
  \renewcommand*\tablename{Table}
\else
  \newcommand\tablename{Table}
\fi
}
\@ifpackageloaded{float}{}{\usepackage{float}}
\floatstyle{ruled}
\@ifundefined{c@chapter}{\newfloat{codelisting}{h}{lop}}{\newfloat{codelisting}{h}{lop}[chapter]}
\floatname{codelisting}{Listing}

\makeatother
\makeatletter
\@ifpackageloaded{caption}{}{\usepackage{caption}}
\@ifpackageloaded{subcaption}{}{\usepackage{subcaption}}
\makeatother
\usepackage{bookmark}
\IfFileExists{xurl.sty}{\usepackage{xurl}}{} 
\hypersetup{
  pdftitle={Machine Learning and Time-Series Foundation Models for Day-Ahead Electricity Forecasting in a University Building},
  pdfauthor={Amey N. Vasulkar; Samuel Kernan Freire; Arie-Willem de Leeuw; Tadeo-Baldiri Salcedo-Rahola},
  pdfkeywords={electricity consumption, day-ahead
forecasting, university buildings, time-series forecasting, time-series
foundation models, workdays and holidays},
  colorlinks=true,
  linkcolor={blue},
  filecolor={Maroon},
  citecolor={Blue},
  urlcolor={Blue},
  pdfcreator={LaTeX via pandoc}}

\title{Machine Learning and Time-Series Foundation Models for Day-Ahead
Electricity Forecasting in a University Building}
\author{Amey N. Vasulkar \and Samuel Kernan Freire \and Arie-Willem de
Leeuw \and Tadeo-Baldiri Salcedo-Rahola}
\date{2026-10-08}
\begin{document}
\maketitle

\section{Abstract}\label{abstract}

Day-ahead electricity forecasts can help building operators schedule
heating, ventilation, and other services around anticipated demand.
Choosing a forecasting model also requires balancing accuracy with its
forecast-time inputs, local training, and computational requirements.

We benchmark statistical, machine-learning, recurrent neural-network,
and pretrained TimesFM models for a university building. We use
chronological validation and evaluate all configurations on the same 57
test days, comparing accuracy across workdays, weekends, and holidays
alongside local fitting and forecasting time. Weather covariates use
retrospective observations.

TimesFM 3.0 with covariates achieves the lowest overall errors, reducing
mean absolute error by approximately 43\% relative to weekly
seasonal-naive forecasting. XGBoost remains competitive, although paired
daily comparisons do not establish a clear mean-absolute-error advantage
for TimesFM. TimesFM also performs strongly using only past electricity
consumption. Adding covariates improves its holiday forecasts but
slightly worsens workday and weekend forecasts, leaving overall mean
absolute error almost unchanged. TimesFM requires no building-specific
weight training, while XGBoost offers rapid local fitting and inference.
The evaluated LSTM takes longer to fit without outperforming either
approach.

An exploratory historical evaluation reveals seasonal variation, while a
separate weather-source comparison produces only small changes in the
selected TimesFM configuration's accuracy. These results support
pretrained TimesFM as a strong initial forecasting baseline for this
building, with XGBoost as a competitive locally fitted alternative.
Covariates are particularly useful when improved holiday forecasts
justify the additional input preparation.

\section{Introduction}\label{introduction}

Buildings have been estimated to account for 36\% of global energy use
and 40\% of associated carbon dioxide emissions (Khalil et al., 2022).
Improving their operation complements improvements to the building
fabric and installed equipment. In particular, scheduling heating,
ventilation, and air conditioning (HVAC) and other building services
around anticipated demand provides an opportunity to align operation
with building activity. Reviews of data-driven building energy modelling
identify forecasting as a component of such management and control
(Bourdeau et al., 2019; Khalil et al., 2022).

Day-ahead electricity forecasting supplies the next 24 hourly
electricity-consumption values before the forecast day begins. The
hourly profile can inform service scheduling and identify periods of
expected high demand, whereas a daily total cannot distinguish when that
demand occurs. Producing such forecasts routinely requires a model that
can capture changing consumption patterns from the available data.
Building operators must therefore choose both a forecasting method and a
procedure for preparing its inputs and keeping it up to date. The
practical value of that choice depends on forecast accuracy, covariate
requirements, local fitting, and inference resources. Statistical
models, tree ensembles, and recurrent neural networks offer different
combinations of these requirements (Khalil et al., 2022). Pretrained
time-series foundation models introduce another possibility by
forecasting new series without building-specific weight training (Das et
al., 2024). Whether that transfer provides a useful alternative to a
well-tuned local model remains an empirical question for the building
concerned.

University buildings provide an informative setting for this comparison
because teaching days, weekends, examinations, and academic holidays
produce changes in operating schedules. Studies of academic calendars,
occupancy, and campus electricity consumption motivate assessing these
periods separately (Alfalah et al., 2023; Ghanavati et al., 2023; Gui et
al., 2021). A forecast can reproduce routine teaching-day variation
while still overstating consumption during a closure. These schedule
changes make average performance across all hours an incomplete basis
for choosing a model. In this paper, we examine a modern, well-insulated
teaching building at the Delft campus of The Hague University of Applied
Sciences (THUAS), extending our preceding study of the same building
(Stienstra et al., 2026).

For models using weather covariates, practical application also requires
a suitable weather source. Historical evaluations can use station
observations or reanalysis, while next-day weather must be supplied by a
forecast issued beforehand. Differences between these products may
affect the electricity-consumption forecast, making weather-source
sensitivity relevant alongside model accuracy.

This study contributes a temporally validated comparison of locally
fitted and pretrained forecasting configurations, together with evidence
on when their accuracy differences matter. We compare models on the same
forecast days, assess differences across workdays and closures, and
quantify uncertainty in their mean daily errors. An extended historical
TimesFM evaluation examines seasonal behaviour beyond the
winter-weighted test; a controlled weather-source comparison examines
whether its forecasts depend strongly on the supplied weather product.
Together with measured fitting and forecast times, these analyses
support model selection according to accuracy, input requirements, and
local modelling effort. Our target is electricity consumption:
tariff-dependent costs and savings from subsequent control decisions are
not evaluated.

\section{Related Work}\label{related-work}

The reviews by Bourdeau et al. (2019) and Khalil et al. (2022) document
a broad range of statistical, machine-learning, and deep-learning
approaches to building energy forecasting. They organise the evidence by
building characteristics, available data, preprocessing, forecast
horizon, and evaluation method. These distinctions are important when
interpreting reported performance: a method evaluated on daily
consumption for one building is not directly comparable with a method
predicting the hourly profile of another. The literature therefore
motivates comparison under explicit experimental conditions rather than
selection of an algorithm solely from a published accuracy value.

University-building studies illustrate the importance of both weather
and operating schedules. Amber et al. (2017) examined five years of
daily electricity consumption for two buildings at London South Bank
University, using multiple regression with weather, weekday, and
building-type covariates. Their 2011 evaluation reported normalised RMSE
of 12\% for the administrative building and 13\% for the academic
building; weekday index, temperature, and building type were significant
predictors. This provides an applied precedent for a comparatively
simple model with contextual covariates. However, its daily resolution
does not establish performance for the 24-hour profile required by
day-ahead scheduling. Studies of academic calendars and occupancy
further motivate distinguishing teaching activity from closures, without
implying that realised attendance is known in advance (Alfalah et al.,
2023; Gui et al., 2021).

Our preceding THUAS study provides a directly relevant comparison of
multiple linear regression, Holt--Winters, Prophet, XGBoost, and LSTM
for the same building (Stienstra et al., 2026). Its reported results
cover 51 workdays: Prophet had MAE 17.78 kWh and \(R^2=0.92\), compared
with MAE 22.53 kWh for XGBoost and 29.21 kWh for LSTM. XGBoost
nevertheless had the lowest MAPE and mean squared error. Thus, even
within one study, the preferred model depended on the error measure, and
the recurrent neural network did not provide the best reported accuracy.
The study also identified difficulties on non-workdays, when lower
consumption and different temporal patterns complicated forecasting. Its
workday-only results leave open how the comparison changes when closures
are included and how performance varies over a longer seasonal record.

Hybrid approaches combine models to improve different parts of a
forecast. For day-ahead forecasting in a German energy community,
Semmelmann et al. (2022) used a bidirectional LSTM to predict the
general consumption pattern and XGBoost to improve peak timing and
magnitude. Adding the peak correction reduced MAPE by 0.14 percentage
points overall and 3.55 percentage points for peaks. Furthermore, the
larger peak-specific gain also shows why an overall score can hide
improvements during the periods a model component is designed to
address. In university buildings, the preceding THUAS study identifies
closures as another period requiring separate attention (Stienstra et
al., 2026). These studies motivate reporting errors for distinct
operating periods alongside the pooled score.

Pretrained forecasting studies introduce a different route to accuracy.
Das et al. (2024) reported zero-shot TimesFM performance approaching
supervised, dataset-specific methods across public benchmark
collections. In an applied electricity-demand comparison,
Llanes-Guilarte et al. (2026) evaluated TimesFM 2.0 and Lag-Llama on
small-community consumption data and reported that their relative
performance depended on temporal resolution. These studies support
investigating pretrained transfer, but they do not establish its
accuracy for a particular university building or for later TimesFM
versions. Building-level evaluation must still examine context
requirements, the value of external covariates, and performance during
schedule changes.

The remaining gap is a clear account of how locally fitted and
pretrained methods compare when the building, forecast hours, and
model-selection procedure are held consistent. Their performance across
routine operation, closures, and seasons also needs to be considered
alongside the effort of preparing covariates and fitting models. Weather
introduces an additional uncertainty: the station observations or
reanalysis used in a historical evaluation differ from the forecasts
available for next-day operation. Establishing how strongly those source
differences affect electricity-consumption forecasts would make the
model comparison more useful in practice.

\section{Research Questions}\label{research-questions}

We address four questions:

\begin{enumerate}
\def\labelenumi{\arabic{enumi}.}
\tightlist
\item
  How do the selected statistical, machine-learning, LSTM, and TimesFM
  configurations compare with daily and weekly seasonal-naive forecasts
  on the same 24-hour targets?
\item
  Are differences in forecast accuracy consistent across workdays,
  weekends, and holidays, and what does an extended historical
  evaluation of TimesFM reveal about seasonal performance?
\item
  How sensitive is a fixed TimesFM 3.0 model with covariates to KNMI
  observations (Royal Netherlands Meteorological Institute, n.d.), ERA5
  reanalysis (Hersbach et al., 2020), and historical ECMWF IFS weather
  (Open-Meteo, n.d.-a)?
\item
  What local fitting time, forecast time, memory use, and historical
  context are required by the evaluated configurations?
\end{enumerate}

The comparison among all models uses the same 57-day test set. Two
additional analyses address the interpretation of this winter-weighted
benchmark: the extended historical evaluation examines TimesFM across
more seasons and closure periods, while the weather-source comparison
tests its sensitivity to the supplied weather product. Neither analysis
changes the selected model settings. We report the resource quantities
measured during execution; total pretraining cost and complete
model-storage costs are outside this comparison.

\section{Methods and Data}\label{methods-and-data}

\subsection{Building, Electricity Consumption, and Time
Convention}\label{building-electricity-consumption-and-time-convention}

We use hourly whole-building electricity consumption for the THUAS Delft
campus building. The institutional dataset was assembled for the
preceding site study, which identifies electricity measurements from the
Joulz portal and scheduling data from THUAS's planning systems
(Stienstra et al., 2026). Electricity consumption is the response
variable, measured in kWh for each hourly interval.

The dataset contains 13,752 hourly observations arranged in 573
consecutive 24-hour forecast windows. It was supplied as a
training/validation file with 12,384 observations and a test file with
1,368 observations. Both files contain a timestamp, electricity
consumption, and 47 candidate covariates. We use the source files
together for descriptive statistics while preserving their chronological
separation for model selection and testing.

All timestamps are reported in UTC. The complete series spans 4
September 2022 at 23:00 UTC through 30 March 2024 at 22:00 UTC;
forecasts are issued daily at 23:00 UTC for the next 24 hourly targets.
We retain the institutional operating-day labels for calendar
comparisons. Appendix Section~\ref{sec-time-weather-alignment} documents
the time conversion and weather alignment.

\begin{figure}

\centering{

\includegraphics[width=1\linewidth,height=\textheight,keepaspectratio,alt={Complete hourly electricity-consumption series with labelled validation and test periods, followed by an expanded view of the first complete UTC week.}]{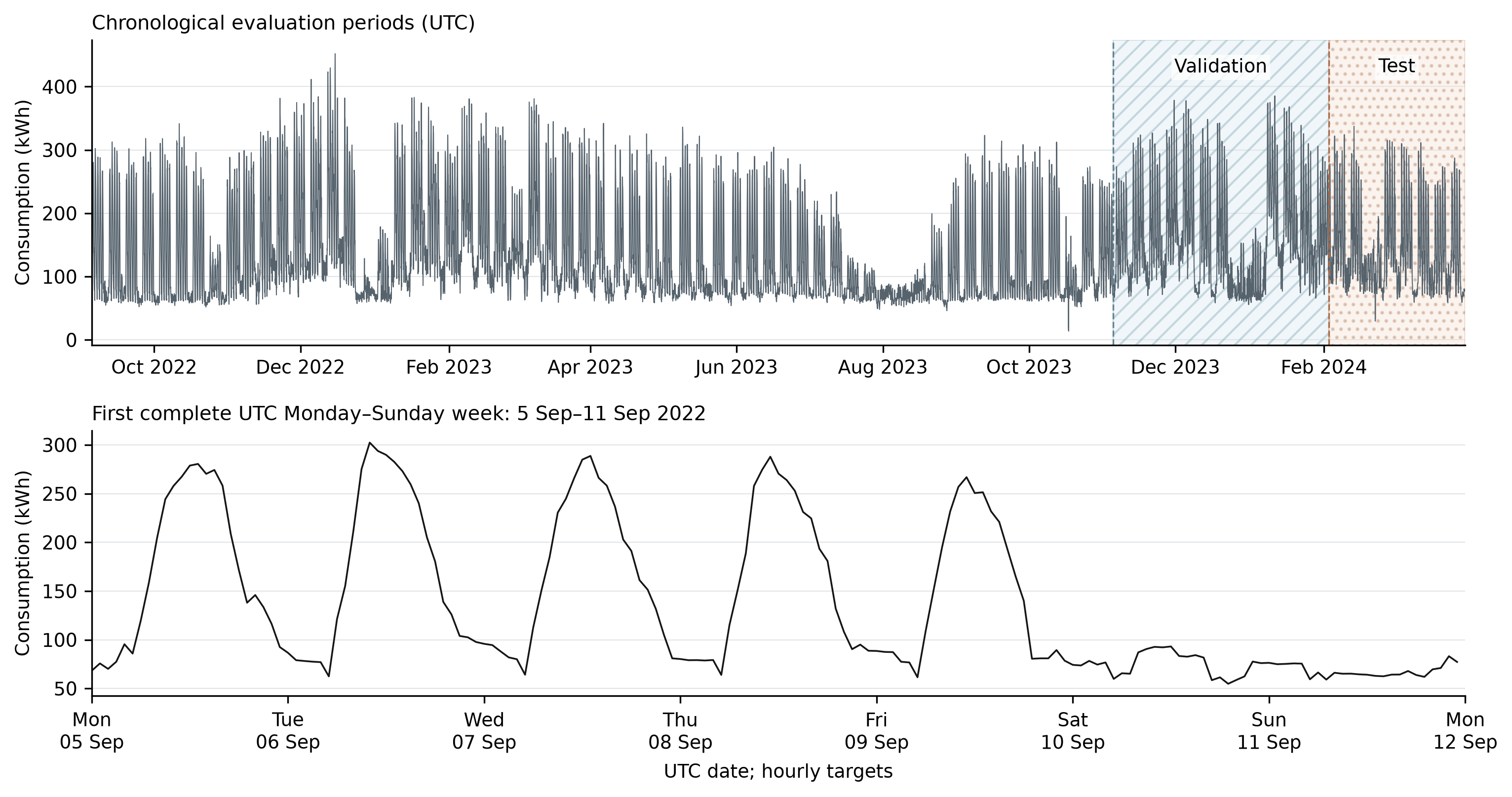}

}

\caption{\label{fig-full-history-load}Whole-building hourly electricity
consumption from 4 September 2022 at 23:00 UTC to 30 March 2024 at 22:00
UTC. The upper panel marks the validation and test periods; the lower
panel expands the first complete Monday--Sunday week in UTC to show the
hourly variation. Both panels retain the original hourly resolution.}

\end{figure}%

\subsection{Weather Data}\label{sec-weather-data}

The main benchmark uses hourly KNMI observations from Rotterdam Airport
station 344, the station included in the supplied building dataset
(Royal Netherlands Meteorological Institute, n.d.). The preceding site
study describes the selected station as approximately 4 km from the
building (Stienstra et al., 2026). These observations provide a regional
weather reference rather than measurements at the building itself. To
examine whether the choice of this reference affects
electricity-consumption forecasts, we also use ERA5 reanalysis and
historical ECMWF IFS forecasts obtained through Open-Meteo (Hersbach et
al., 2020; Open-Meteo, n.d.-b, n.d.-a). Table~\ref{tbl-weather-sources}
distinguishes their roles.

\begin{longtable}[]{@{}
  >{\raggedright\arraybackslash}p{(\linewidth - 4\tabcolsep) * \real{0.3333}}
  >{\raggedright\arraybackslash}p{(\linewidth - 4\tabcolsep) * \real{0.3333}}
  >{\raggedright\arraybackslash}p{(\linewidth - 4\tabcolsep) * \real{0.3333}}@{}}
\caption{Weather sources and their role. The main 11-model benchmark
uses KNMI for weather covariates; the additional source comparison uses
a fixed TimesFM 3.0
configuration.}\label{tbl-weather-sources}\tabularnewline
\toprule\noalign{}
\begin{minipage}[b]{\linewidth}\raggedright
Source
\end{minipage} & \begin{minipage}[b]{\linewidth}\raggedright
Product
\end{minipage} & \begin{minipage}[b]{\linewidth}\raggedright
Role in the comparison
\end{minipage} \\
\midrule\noalign{}
\endfirsthead
\toprule\noalign{}
\begin{minipage}[b]{\linewidth}\raggedright
Source
\end{minipage} & \begin{minipage}[b]{\linewidth}\raggedright
Product
\end{minipage} & \begin{minipage}[b]{\linewidth}\raggedright
Role in the comparison
\end{minipage} \\
\midrule\noalign{}
\endhead
\bottomrule\noalign{}
\endlastfoot
KNMI & Observations at Rotterdam Airport station 344 & Observed-weather
reference \\
ERA5 & Gridded reanalysis & Retrospective weather reconstruction \\
Historical IFS & Archived ECMWF IFS forecasts from Open-Meteo &
Forecast-product substitution \\
\end{longtable}

The supplied dataset contains 21 candidate weather quantities. We retain
the seven with corresponding definitions across all three products and
harmonise their units: air temperature, dew-point temperature, relative
humidity, mean-sea-level pressure, hourly precipitation, wind speed, and
wind direction. The other 14 weather quantities are excluded, as
described in the next section. The retained quantities' spatial and
temporal sampling are not identical; Appendix
Section~\ref{sec-time-weather-alignment} documents the alignment and
remaining differences. Reanalysis describes past conditions using
observations and a numerical model; it is not a weather forecast issued
before the target day. Historical IFS is an archived forecast product,
although the downloaded records do not identify when each individual
value became available. These distinctions let us test weather-source
sensitivity without assuming that all three products were accessible at
the historical forecast origin.

\subsection{Forecast-Time Inputs}\label{forecast-time-inputs}

We reduce the 47 candidate covariates to 30 in two steps. First, we
remove 14 of the 21 candidate weather quantities because they lack
consistently defined counterparts across the three sources, retaining
the seven listed in Section~\ref{sec-weather-data}. The removed
quantities include weather codes and additional daily aggregates.
Second, we exclude three further covariates: realised people counts,
cumulative room bookings, and the one-hour electricity-consumption lag,
whose target-period values are unavailable for the complete day-ahead
forecast. Thus, \(47-14-3=30\) covariates remain: 21 calendar and
schedule quantities, seven weather quantities, and two historical
electricity-consumption lags (Table~\ref{tbl-feature-availability}).
Models supporting external covariates use these same quantities;
seasonal-naive forecasts and target-only LSTM and TimesFM models use
past electricity consumption, while Prophet base uses its trend,
seasonal, and holiday components.

\begin{longtable}[]{@{}
  >{\raggedright\arraybackslash}p{(\linewidth - 6\tabcolsep) * \real{0.2000}}
  >{\raggedright\arraybackslash}p{(\linewidth - 6\tabcolsep) * \real{0.4000}}
  >{\raggedright\arraybackslash}p{(\linewidth - 6\tabcolsep) * \real{0.1500}}
  >{\raggedright\arraybackslash}p{(\linewidth - 6\tabcolsep) * \real{0.2500}}@{}}
\caption{The 30 retained covariates, grouped into three domains.
Availability refers to the forecast origin. Appendix
Section~\ref{sec-covariate-catalogue} gives the individual stored
variables.}\label{tbl-feature-availability}\tabularnewline
\toprule\noalign{}
\begin{minipage}[b]{\linewidth}\raggedright
Domain
\end{minipage} & \begin{minipage}[b]{\linewidth}\raggedright
Covariates (number)
\end{minipage} & \begin{minipage}[b]{\linewidth}\raggedright
Units
\end{minipage} & \begin{minipage}[b]{\linewidth}\raggedright
Availability
\end{minipage} \\
\midrule\noalign{}
\endfirsthead
\toprule\noalign{}
\begin{minipage}[b]{\linewidth}\raggedright
Domain
\end{minipage} & \begin{minipage}[b]{\linewidth}\raggedright
Covariates (number)
\end{minipage} & \begin{minipage}[b]{\linewidth}\raggedright
Units
\end{minipage} & \begin{minipage}[b]{\linewidth}\raggedright
Availability
\end{minipage} \\
\midrule\noalign{}
\endhead
\bottomrule\noalign{}
\endlastfoot
Calendar and schedule (21) & Hour, year, month, week, day, day of week
(6) & Calendar indices & Known in advance \\
& Sine/cosine pairs for hour, day, day of week, week, month (10) &
Dimensionless & Known in advance \\
& Work-week, weekend, holiday, opening-hours, examination-week
indicators (5) & Binary & Calendar known; schedules assumed known \\
Weather (7) & Air and dew-point temperatures (2) & °C & Retrospective \\
& Relative humidity; sea-level pressure (2) & \%; hPa & Retrospective \\
& Hourly precipitation (1) & mm & Retrospective \\
& Wind speed; wind direction (2) & m/s; degrees & Retrospective \\
Historical consumption (2) & Electricity consumption 24 and 168 hours
earlier (2) & kWh & Observed before the forecast origin \\
\end{longtable}

We assume that the supplied opening-hours and examination schedules were
available before the forecast day; their historical publication times
were not retained. Realised people counts and cumulative room bookings
describe activity during the target period. The one-hour lag would
require unobserved consumption after the first forecast step, whereas
the 24- and 168-hour lags are available for every target hour.

\subsection{Models}\label{models}

The benchmark contains 11 configurations, listed in
Table~\ref{tbl-model-design}. Ridge is multiple linear regression with a
penalty that shrinks its coefficients (Hoerl \& Kennard, 1970). XGBoost
combines regression trees and can represent nonlinear relationships and
interactions between covariates (Chen \& Guestrin, 2016). Prophet
combines trend, daily and weekly seasonality, and holiday effects; we
test it with and without external covariates (Taylor \& Letham, 2018).
The LSTM uses recurrent memory cells to learn temporal relationships
(Hochreiter \& Schmidhuber, 1997); our implementation receives sequences
of past electricity consumption and produces the next 24 hourly values.
Its hybrid adds a ridge prediction of the LSTM residual.

TimesFM is a decoder-only time-series foundation model pretrained on
collections of real and synthetic series (Das et al., 2024). Zero-shot
forecasting means that its network weights are not fitted to this
building; past electricity consumption is still required as context. We
use the pretrained TimesFM 2.5 and 3.0 weights supplied by Google
Research, retaining 2.5 as a version reference (Google Research, 2025,
2026). TimesFM 3.0, released in August 2026, supports native
multivariate forecasting with historical and future covariates. We
evaluate both versions using target history alone, and additionally
evaluate TimesFM 3.0 with the same 30 covariates used by the
regression-based methods. This native interface differs from the
external ridge residual modelling used in the TimesFM 2.5 hybrids.
Context length and inference settings are selected by validation,
without changing the pretrained weights.

\begin{longtable}[]{@{}
  >{\raggedright\arraybackslash}p{(\linewidth - 4\tabcolsep) * \real{0.3333}}
  >{\raggedright\arraybackslash}p{(\linewidth - 4\tabcolsep) * \real{0.3333}}
  >{\raggedright\arraybackslash}p{(\linewidth - 4\tabcolsep) * \real{0.3333}}@{}}
\caption{Models and their forecast-time inputs. Prophet represents the
holiday indicator through its native holiday component, leaving 29
external regressors.}\label{tbl-model-design}\tabularnewline
\toprule\noalign{}
\begin{minipage}[b]{\linewidth}\raggedright
Configuration
\end{minipage} & \begin{minipage}[b]{\linewidth}\raggedright
Forecast-time inputs
\end{minipage} & \begin{minipage}[b]{\linewidth}\raggedright
Building-specific fitting
\end{minipage} \\
\midrule\noalign{}
\endfirsthead
\toprule\noalign{}
\begin{minipage}[b]{\linewidth}\raggedright
Configuration
\end{minipage} & \begin{minipage}[b]{\linewidth}\raggedright
Forecast-time inputs
\end{minipage} & \begin{minipage}[b]{\linewidth}\raggedright
Building-specific fitting
\end{minipage} \\
\midrule\noalign{}
\endhead
\bottomrule\noalign{}
\endlastfoot
Seasonal-naive 24 h & Previous day's consumption at the corresponding
hour & None \\
Seasonal-naive 168 h & Previous week's consumption at the corresponding
hour & None \\
Ridge & 30 covariates & Standardisation and regression fitted at each
origin \\
XGBoost & 30 covariates & Regression trees fitted at each origin \\
Prophet base & Past consumption and native calendar/holiday components &
Trend, seasonality, and holiday effects fitted at each origin \\
Prophet with regressors & Past consumption, 29 external covariates, and
native holiday effects & Prophet and its regressors fitted jointly at
each origin \\
TimesFM 2.5 target-only & Past consumption & Pretrained weights remain
fixed \\
LSTM base & Past consumption & Network and scaling fitted at each
origin \\
LSTM + ridge & LSTM forecast and 30 covariates for residual correction &
LSTM and residual regression fitted from prior observations \\
TimesFM 3.0 target-only & Past consumption & Pretrained weights remain
fixed \\
TimesFM 3.0 with covariates & Past consumption and 30 covariates &
Pretrained weights remain fixed \\
\end{longtable}

For a forecast origin \(t\), let \(y_{t+h}\in\mathbb{R}\) denote
electricity consumption at target hour \(t+h\), with \(h=0,\ldots,23\).
At each of these 24 target hours, \(\mathbf{x}_{t+h}\in\mathbb{R}^{30}\)
contains the 30 covariates; \(y_{<t}\) denotes past consumption. The
hour index \(h\) and the number of covariates therefore describe
different dimensions. A model \(M\) with parameters \(\theta\) and
external covariates produces the forecast in
Equation~\ref{eq-covariate-forecast}:

\begin{equation}\protect\phantomsection\label{eq-covariate-forecast}{
\hat y_{t+h}=M_{\theta}\!\left(y_{<t},\mathbf{x}_{t+h}\right),
\qquad h=0,\ldots,23.
}\end{equation}

The seasonal-naive forecasts in Equation~\ref{eq-seasonal-naive} copy an
observed value from the corresponding hour one day or one week earlier:

\begin{equation}\protect\phantomsection\label{eq-seasonal-naive}{
\hat y_{t+h}^{(k)}=y_{t+h-k},
\qquad k\in\{24,168\}.
}\end{equation}

For the LSTM hybrid, \(r_j\) in Equation~\ref{eq-residual} is the
observed error of an earlier base forecast at target hour \(j\). Ridge
models that error from the covariates, and its prediction is added to
the LSTM forecast:

\begin{equation}\protect\phantomsection\label{eq-residual}{
r_j=y_j-\hat y^{\mathrm{LSTM}}_j,
\qquad
\hat y^{\mathrm{hybrid}}_{t+h}
=\hat y^{\mathrm{LSTM}}_{t+h}+\hat r_{t+h}.
}\end{equation}

Only residuals whose actual consumption is known before \(t\) enter the
correction fit. The correction is added once. This follows the general
residual-modelling principle of Zhang (2003).

Appendix Section~\ref{sec-timesfm25-hybrids} compares the TimesFM 2.5
target-only model with two jointly tuned residual hybrids: TimesFM
followed by ridge correction, and ridge followed by TimesFM forecasting
of residuals. These additional experiments examine whether external
residual modelling improves on the base forecast. TimesFM 2.5
target-only also remains in the main benchmark as a version reference.

\subsection{Training, Validation, and Test
Design}\label{training-validation-and-test-design}

We use sliding-window, rolling-origin cross-validation, following the
chronological evaluation principle described by Tashman (2000). At each
forecast origin, refitted models use the immediately preceding 8,760
hourly observations, equivalent to 365 days, and predict the following
24 hours. Both ends of the training window then advance by 24 hours.
Scalers and imputers are fitted only to that origin's history. XGBoost
reserves the final 14 days of its training window for early stopping,
while the LSTM uses the final 10\% of its training sequences. Prophet's
yearly seasonality is disabled because each fit contains only one annual
cycle. TimesFM keeps its pretrained weights fixed and receives a context
length selected by validation.

\begin{longtable}[]{@{}
  >{\raggedright\arraybackslash}p{(\linewidth - 8\tabcolsep) * \real{0.1765}}
  >{\raggedright\arraybackslash}p{(\linewidth - 8\tabcolsep) * \real{0.1765}}
  >{\raggedleft\arraybackslash}p{(\linewidth - 8\tabcolsep) * \real{0.2353}}
  >{\raggedleft\arraybackslash}p{(\linewidth - 8\tabcolsep) * \real{0.2353}}
  >{\raggedright\arraybackslash}p{(\linewidth - 8\tabcolsep) * \real{0.1765}}@{}}
\caption{Chronological evaluation periods. Each forecast day begins at
23:00 UTC and contains 24 consecutive
targets.}\label{tbl-split}\tabularnewline
\toprule\noalign{}
\begin{minipage}[b]{\linewidth}\raggedright
Stage
\end{minipage} & \begin{minipage}[b]{\linewidth}\raggedright
Target interval in UTC
\end{minipage} & \begin{minipage}[b]{\linewidth}\raggedleft
Forecast days
\end{minipage} & \begin{minipage}[b]{\linewidth}\raggedleft
Hourly targets
\end{minipage} & \begin{minipage}[b]{\linewidth}\raggedright
Purpose
\end{minipage} \\
\midrule\noalign{}
\endfirsthead
\toprule\noalign{}
\begin{minipage}[b]{\linewidth}\raggedright
Stage
\end{minipage} & \begin{minipage}[b]{\linewidth}\raggedright
Target interval in UTC
\end{minipage} & \begin{minipage}[b]{\linewidth}\raggedleft
Forecast days
\end{minipage} & \begin{minipage}[b]{\linewidth}\raggedleft
Hourly targets
\end{minipage} & \begin{minipage}[b]{\linewidth}\raggedright
Purpose
\end{minipage} \\
\midrule\noalign{}
\endhead
\bottomrule\noalign{}
\endlastfoot
Initial history & 4 September 2022, 23:00--4 November 2023, 22:00 & 426
& 10,224 & Supply past observations for subsequent forecasts \\
Validation & 4 November 2023, 23:00--2 February 2024, 22:00 & 90 & 2,160
& Select hyperparameters \\
Test & 2 February 2024, 23:00--30 March 2024, 22:00 & 57 & 1,368 &
Evaluate the selected configurations \\
\end{longtable}

A hyperparameter candidate is one complete combination of settings. Each
candidate is evaluated over the same 90 consecutive daily folds,
producing 2,160 validation predictions. We select the lowest pooled
RMSE, using MAE and then candidate order to break ties. We then hold
those settings fixed during the 57-day test and continue advancing the
forecast origin by one day; models whose policy requires daily fitting
are refitted on the available history. The test is not used to tune
these settings.

Ridge uses a grid of 25 log-spaced penalties. Optuna's tree-structured
Parzen estimator proposes 96 trials for XGBoost, 96 for each Prophet
configuration, 72 for the LSTM base, and 48 for its ridge residual
correction (Akiba et al., 2019). All five Optuna searches use seed
20260729 and disable trial pruning, so every candidate is scored over
the complete 90-day validation period. This differs from early stopping
within an individual XGBoost or LSTM fit, which limits fitting
iterations without dropping forecast days. TimesFM 2.5 uses a grid of
eight combinations of context length and input normalisation. Each
TimesFM 3.0 configuration uses a 64-combination grid: eight context
lengths from 168 to 8,760 hours, symmetric averaging on/off, a
non-negative-output option on/off, and input z-normalisation on/off. The
forecast horizon remains 24 hours throughout.

The selected TimesFM 2.5 context is 672 hours. Both TimesFM 3.0
configurations select 8,760 hours, symmetric averaging, no
non-negative-output transform, and no optional z-normalisation; the
point prediction is the median forecast quantile. The selected LSTM uses
a 168-hour lookback and one 64-unit layer. Complete validation scores
and selected settings are reported in Appendix
Section~\ref{sec-validation-settings}.

\subsection{Descriptive Analysis}\label{descriptive-analysis}

We summarise the distribution of hourly electricity consumption, mean
hourly profiles, and differences between workdays, weekends, and
holidays. A workday is a day labelled as neither a holiday nor a
weekend; a holiday label takes precedence over a weekend label when
forming the three separate groups. We report counts with every
comparison.

Pearson correlations describe the direction and strength of linear
associations between consumption and individual covariates. Full-period
correlations characterise the dataset; they are not used to select
models from the test outcomes. Daily and weekly autocorrelation and
stationarity diagnostics provide supporting descriptions in the
appendix. The main Results section focuses on consumption levels,
calendar patterns, and the three covariate domains.

\subsection{Additional TimesFM
Analyses}\label{additional-timesfm-analyses}

We use two supplementary analyses to examine conditions not resolved by
the main model comparison. The extended historical evaluation asks how
forecast errors change across seasons and closure periods beyond the
short winter test. The weather-source comparison asks whether a TimesFM
forecast changes materially when the weather description changes. Both
use the selected TimesFM 3.0 configuration with covariates; neither
involves further hyperparameter selection or building-specific weight
training.

For the historical evaluation, we start with 168 hours of context at 11
September 2022, 23:00 UTC, and advance the forecast origin by 24 hours
for 566 daily forecasts. Context grows by 24 hours per origin until it
reaches 8,760 hours, after which the window slides at that fixed length.
Every forecast covers only the next 24 hours and receives only earlier
electricity-consumption observations. The pretrained model can therefore
be examined during the first year of this short record without reserving
that entire year for local weight fitting. The weekly seasonal-naive
forecast uses the corresponding hours from the preceding week and
provides a reference for changing forecasting difficulty. This is an
advantage of the selected pretrained evaluation procedure, not a claim
that shorter-history evaluation is impossible with other methods.

We aggregate the historical forecast errors by workday, weekend,
holiday, UTC calendar month, and meteorological season. Winter comprises
December--February, spring March--May, summer June--August, and autumn
September--November. Monthly and seasonal groups follow target
timestamps and may divide a forecast window; we report their hourly
counts and context ranges. Because the fixed settings were selected
using later validation data, this extended historical evaluation is
exploratory rather than an independent all-season test.

The weather comparison uses the three products described in
Section~\ref{sec-weather-data}. It keeps electricity-consumption
history, the other 23 covariates, forecast windows, and all TimesFM 3.0
settings fixed, replacing only the seven weather covariates in both the
8,760-hour context and target hours. We harmonise units, use
mean-sea-level pressure throughout, and assign weather to matching
hourly intervals in UTC. This tests substitution of one complete weather
source for another; it does not test removing weather or independently
changing only the future weather values. Appendix
Section~\ref{sec-time-weather-alignment} describes the interval mapping
and treatment of missing wind directions.

This comparison uses an explicit interval assignment across all three
weather sources. Its KNMI forecasts are therefore reported separately
from the original benchmark, which retains the supplied weather join.
The archived IFS data allow a forecast-product comparison, but the
absence of per-value issue times prevents reconstructing exactly which
forecast an operator could have obtained at each historical origin.

\subsection{Accuracy and Day-Level
Uncertainty}\label{accuracy-and-day-level-uncertainty}

We compute pooled hourly MAE, RMSE, \(R^2\), MAPE, and sMAPE, combining
absolute and percentage error measures to describe forecast accuracy
(Hyndman \& Koehler, 2006). For actual consumption \(y_i\), forecast
\(\hat y_i\), error \(e_i=y_i-\hat y_i\), \(n\) target hours, and mean
observed consumption \(\bar y\), Equation~\ref{eq-metrics} defines these
measures:

\begin{equation}\protect\phantomsection\label{eq-metrics}{
\begin{aligned}
\mathrm{MAE}&=\frac{1}{n}\sum_i |e_i|,&
\mathrm{RMSE}&=\sqrt{\frac{1}{n}\sum_i e_i^2},\\
R^2&=1-\frac{\sum_i e_i^2}{\sum_i(y_i-\bar y)^2},&
\mathrm{MAPE}&=\frac{100}{n}\sum_i\left|\frac{e_i}{y_i}\right|,\\
\mathrm{sMAPE}&=\frac{200}{n}\sum_i\frac{|e_i|}{|y_i|+|\hat y_i|}.&&
\end{aligned}
}\end{equation}

MAE gives the average size of an hourly error in kWh. RMSE gives larger
errors more weight because it squares errors before averaging. Reporting
both reveals the influence of large misses on overall accuracy. Median
absolute error would describe the middle error but would give little
information about that upper tail. MAPE expresses errors relative to
measured consumption; its denominator is nonzero throughout this
dataset, whose minimum consumption is 13.9 kWh.

We calculate \(R^2\) from the pooled target hours of the reported group,
not by averaging daily \(R^2\) values. Its denominator measures
variation around that group's mean. During a relatively flat closure
period, even modest errors can therefore produce a low or negative
\(R^2\); for a constant series it is undefined. A negative value means
that squared forecast error exceeds the error of the group's mean-value
reference, not that consumption has no temporal pattern. We interpret
\(R^2\) alongside MAE and RMSE, which retain a directly interpretable
hourly kWh scale across workdays and closures (Hyndman \& Koehler,
2006).

To assess whether the pooled differences depend on a few forecast days,
we calculate a daily MAE from each 24-hour forecast and compare models
on the same days. XGBoost provides a common fitted-model reference; this
choice sets the direction of the reported differences and does not
determine which model wins. We also compare the TimesFM versions
directly. For model \(m\) and reference \(q\), the daily difference in
Equation~\ref{eq-daily-difference} is negative when model \(m\) is more
accurate:

\begin{equation}\protect\phantomsection\label{eq-daily-difference}{
d_d^{(m,q)}
=\frac{1}{24}\sum_{h=0}^{23}
\left(
|y_{d,h}-\hat y^{(m)}_{d,h}|
-|y_{d,h}-\hat y^{(q)}_{d,h}|
\right).
}\end{equation}

We resample the 57 paired days with replacement 10,000 times, using seed
20260727, and report the 2.5th and 97.5th percentiles of the mean
difference as a percentile bootstrap interval (Davison \& Hinkley,
1997). Keeping each day together preserves dependence among its 24
hourly errors. We apply the same procedure to the three pairwise
weather-source comparisons, using the fixed TimesFM 3.0 configuration
throughout. An interval containing zero indicates that these days do not
clearly establish the direction of the mean difference; it does not
demonstrate equal performance. The intervals are not adjusted for
multiple comparisons and do not preserve dependence between successive
days. Day-type scores are additional descriptive comparisons and do not
select hyperparameters.

To illustrate the hourly forecast profiles, we select the first
chronological test forecast origin in each of the workday, weekend, and
holiday groups. We display the observed consumption and saved forecasts
from TimesFM 3.0 with and without covariates and XGBoost for those
windows. This rule is independent of forecast errors; the aggregate
comparisons use all 57 test days.

\subsection{Computational
Measurements}\label{computational-measurements}

We measure fitting or updating time separately from the time needed by
an already-loaded model to forecast 24 hours. We report the median and
95th percentile across test origins, peak process memory, and peak GPU
memory where recorded. For the LSTM hybrid, the correction time is
reported separately from the LSTM training time it requires. These
measurements describe the executed configurations on their recorded
hardware. They exclude the foundation model's original pretraining cost
and are not a comparison of total model-development costs.

\section{Results}\label{results}

\subsection{Electricity Consumption and Calendar
Patterns}\label{electricity-consumption-and-calendar-patterns}

The dataset has no missing or duplicated hourly timestamps and no
missing or negative electricity-consumption values. Mean hourly
consumption is 140.31 kWh, the median is 112.15 kWh, and values range
from 13.9 to 452.0 kWh (Table~\ref{tbl-descriptive}). The time series in
Figure~\ref{fig-full-history-load} shows changes across the study
period, while the expanded week makes the repeated daytime rises and
overnight consumption visible. The marked validation and test periods
locate the model comparison within this longer record.

\begin{longtable}[]{@{}
  >{\raggedright\arraybackslash}p{(\linewidth - 12\tabcolsep) * \real{0.1429}}
  >{\raggedright\arraybackslash}p{(\linewidth - 12\tabcolsep) * \real{0.1429}}
  >{\raggedright\arraybackslash}p{(\linewidth - 12\tabcolsep) * \real{0.1429}}
  >{\raggedright\arraybackslash}p{(\linewidth - 12\tabcolsep) * \real{0.1429}}
  >{\raggedright\arraybackslash}p{(\linewidth - 12\tabcolsep) * \real{0.1429}}
  >{\raggedright\arraybackslash}p{(\linewidth - 12\tabcolsep) * \real{0.1429}}
  >{\raggedright\arraybackslash}p{(\linewidth - 12\tabcolsep) * \real{0.1429}}@{}}
\caption{Electricity-consumption summary over the complete dataset.
Consumption values are kWh per hourly interval. Workdays exclude
weekends and holidays; holiday labels take
precedence.}\label{tbl-descriptive}\tabularnewline
\toprule\noalign{}
\begin{minipage}[b]{\linewidth}\raggedright
Group
\end{minipage} & \begin{minipage}[b]{\linewidth}\raggedright
Hourly observations
\end{minipage} & \begin{minipage}[b]{\linewidth}\raggedright
Mean (kWh)
\end{minipage} & \begin{minipage}[b]{\linewidth}\raggedright
Median (kWh)
\end{minipage} & \begin{minipage}[b]{\linewidth}\raggedright
SD (kWh)
\end{minipage} & \begin{minipage}[b]{\linewidth}\raggedright
Minimum (kWh)
\end{minipage} & \begin{minipage}[b]{\linewidth}\raggedright
Maximum (kWh)
\end{minipage} \\
\midrule\noalign{}
\endfirsthead
\toprule\noalign{}
\begin{minipage}[b]{\linewidth}\raggedright
Group
\end{minipage} & \begin{minipage}[b]{\linewidth}\raggedright
Hourly observations
\end{minipage} & \begin{minipage}[b]{\linewidth}\raggedright
Mean (kWh)
\end{minipage} & \begin{minipage}[b]{\linewidth}\raggedright
Median (kWh)
\end{minipage} & \begin{minipage}[b]{\linewidth}\raggedright
SD (kWh)
\end{minipage} & \begin{minipage}[b]{\linewidth}\raggedright
Minimum (kWh)
\end{minipage} & \begin{minipage}[b]{\linewidth}\raggedright
Maximum (kWh)
\end{minipage} \\
\midrule\noalign{}
\endhead
\bottomrule\noalign{}
\endlastfoot
Full series & 13752.0 & 140.31 & 112.15 & 76.86 & 13.90 & 452.00 \\
Holiday & 2760 & 87.77 & 77.10 & 33.57 & 13.90 & 374.30 \\
Weekday & 8040 & 176.23 & 161.25 & 79.16 & 56.30 & 452.00 \\
Weekend & 2952 & 91.58 & 83.00 & 30.47 & 46.10 & 306.10 \\
\end{longtable}

\begin{figure}

\centering{

\includegraphics[width=1\linewidth,height=\textheight,keepaspectratio,alt={Directly labelled curves with different line styles show a pronounced workday daytime rise and flatter weekend and holiday consumption profiles.}]{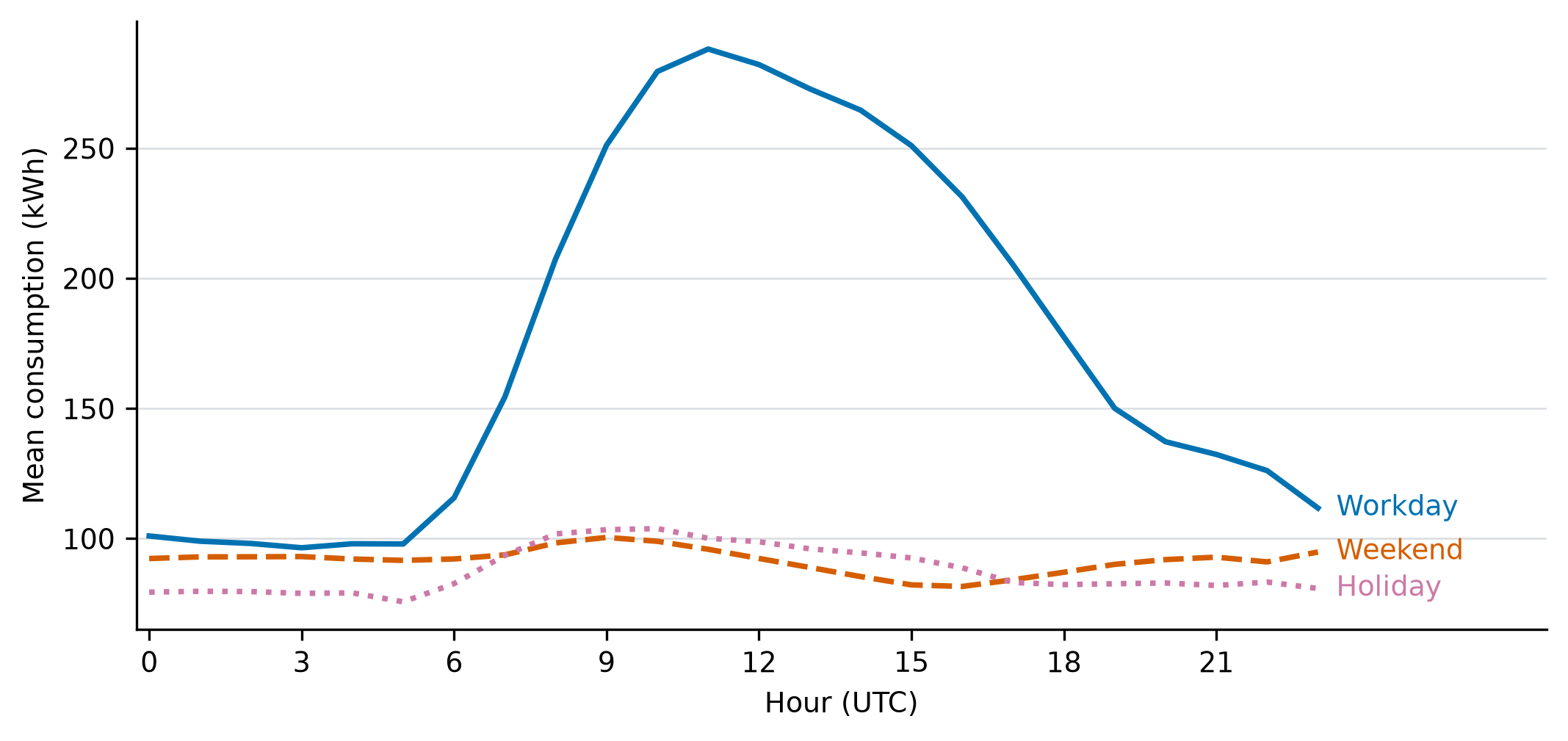}

}

\caption{\label{fig-calendar-regimes}Mean hourly electricity-consumption
profiles for workdays, weekends, and institutional holidays over the
complete dataset. Direct labels and distinct line styles identify the
three groups. Hours are reported in UTC; the original operating-day
memberships are retained.}

\end{figure}

Mean hourly consumption is substantially higher on workdays (176.23 kWh)
than on weekends (91.58 kWh) or holidays (87.77 kWh). The separation is
concentrated in daytime operation (Figure~\ref{fig-calendar-regimes}).

\begin{figure}

\centering{

\includegraphics[width=1\linewidth,height=\textheight,keepaspectratio,alt={Three panels on the same correlation scale show stronger linear associations for opening hours and consumption lags than for the retained weather quantities.}]{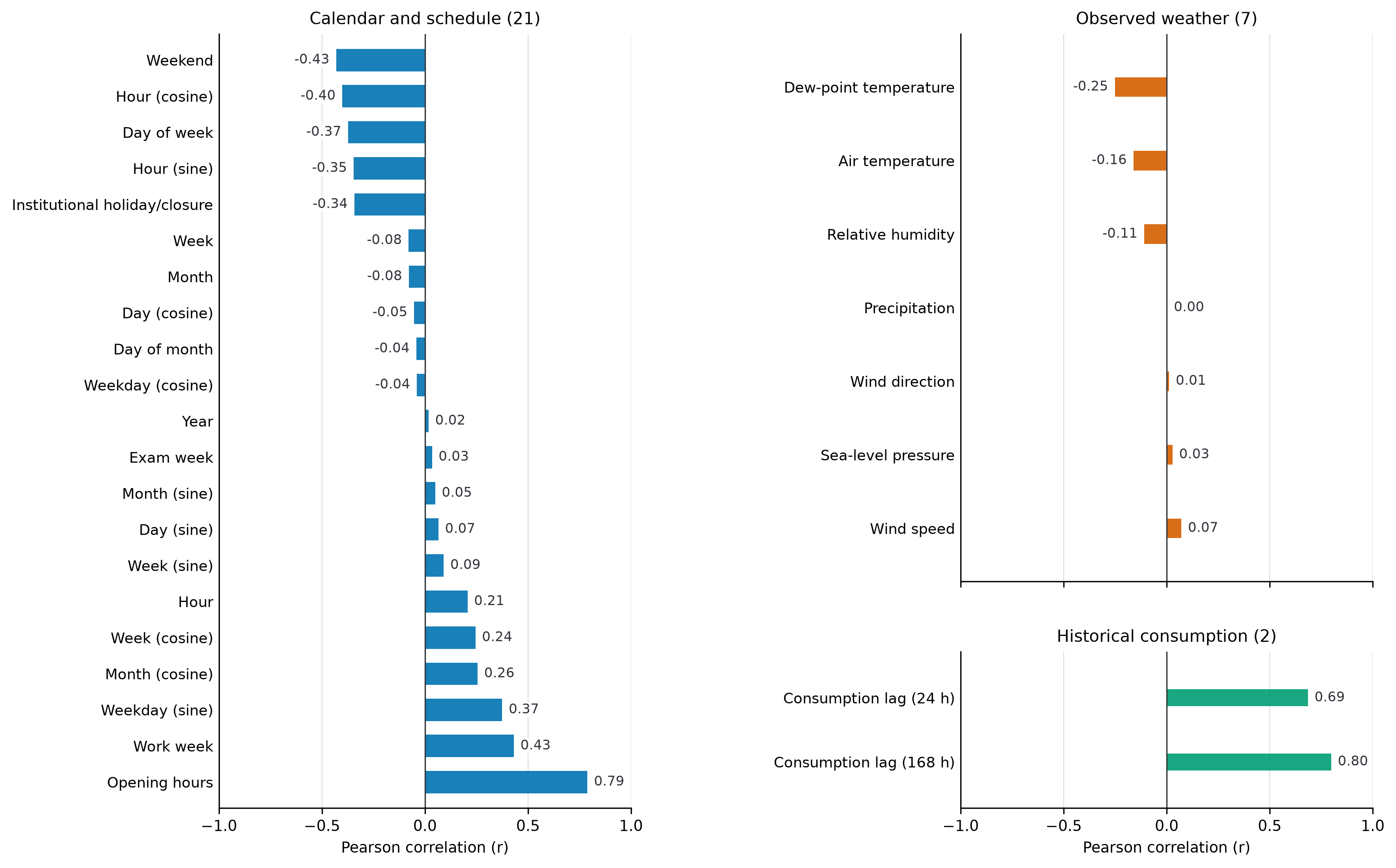}

}

\caption{\label{fig-grouped-correlations}Pearson correlations between
the 30 retained covariates and electricity consumption over the complete
dataset. The calendar/schedule, weather, and historical-consumption
panels use a common correlation scale and consistent bar thickness; the
zero line separates positive and negative associations. Target-day
weather is retrospective.}

\end{figure}

The grouped correlations in Figure~\ref{fig-grouped-correlations} show
the relative associations of calendar, weather, and past-consumption
covariates with electricity consumption. Opening hours and the daily and
weekly lags have strong positive associations, with full-period Pearson
correlations of approximately 0.79, 0.69, and 0.80, respectively. The
daily and weekly autocorrelations are 0.687 and 0.788 over the full
series (Appendix Section~\ref{sec-descriptive-diagnostics}).

\subsection{Benchmark Accuracy}\label{benchmark-accuracy}

\begin{figure}

\centering{

\includegraphics[width=1\linewidth,height=\textheight,keepaspectratio,alt={Two aligned horizontal dot plots compare all 11 configurations, with TimesFM and XGBoost at the lower-error end and consistent model ordering across MAE and RMSE.}]{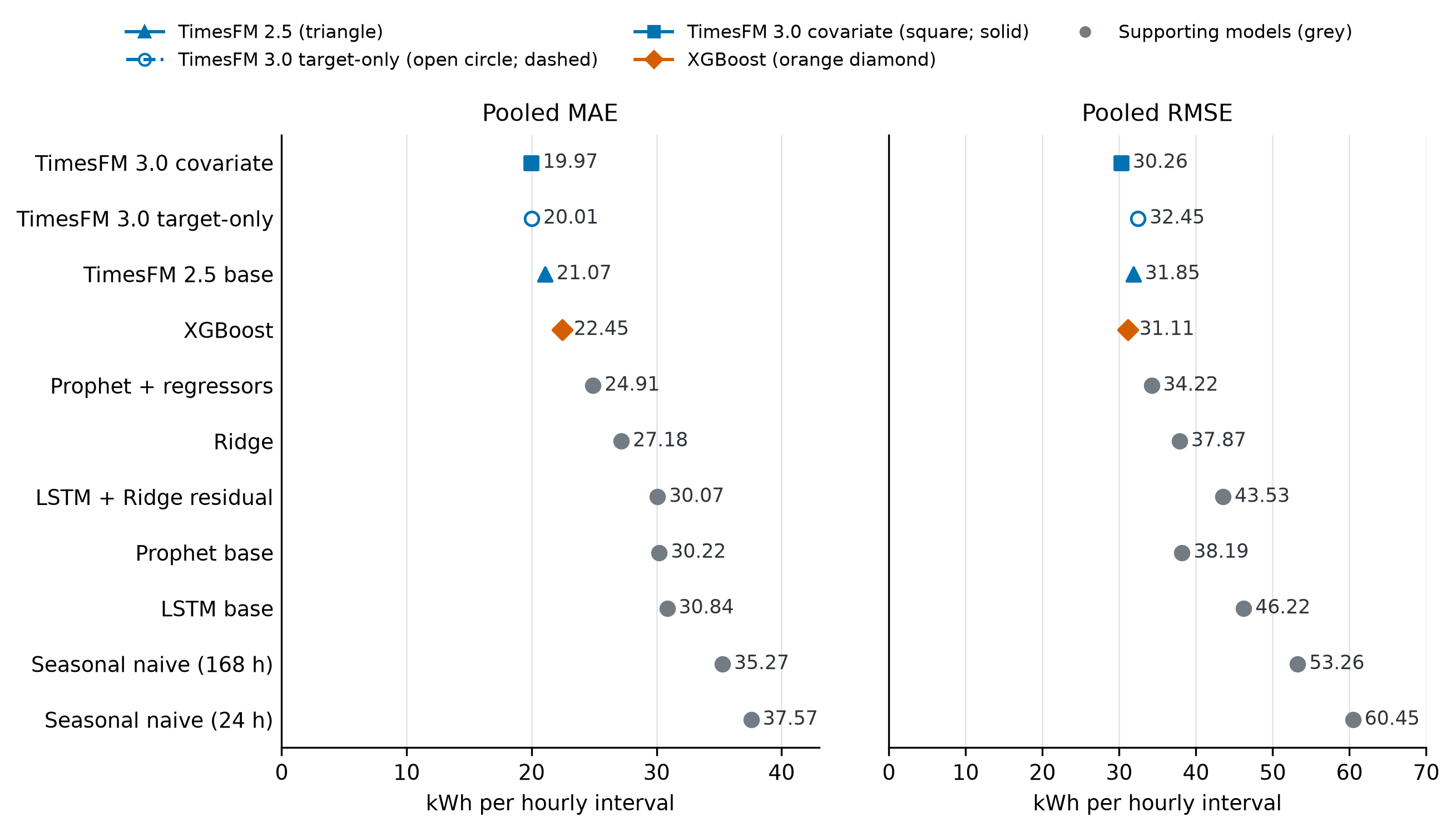}

}

\caption{\label{fig-benchmark-accuracy}Pooled MAE and RMSE on the 1,368
common test targets, with configurations ordered by increasing MAE from
top to bottom in both panels. Points further left indicate lower error.
Colours and marker shapes distinguish TimesFM configurations, XGBoost,
and the supporting comparisons; every configuration is labelled.}

\end{figure}

All 11 configurations in Table~\ref{tbl-model-design} forecast the same
57 test windows and 1,368 hourly targets. Table~\ref{tbl-benchmark}
reports their pooled errors and \(R^2\).
Figure~\ref{fig-benchmark-accuracy} places the configurations in the
same MAE order in both panels, allowing their MAE and RMSE positions to
be compared directly.

\begin{longtable}[]{@{}
  >{\raggedright\arraybackslash}p{(\linewidth - 12\tabcolsep) * \real{0.1429}}
  >{\raggedright\arraybackslash}p{(\linewidth - 12\tabcolsep) * \real{0.1429}}
  >{\raggedright\arraybackslash}p{(\linewidth - 12\tabcolsep) * \real{0.1429}}
  >{\raggedright\arraybackslash}p{(\linewidth - 12\tabcolsep) * \real{0.1429}}
  >{\raggedright\arraybackslash}p{(\linewidth - 12\tabcolsep) * \real{0.1429}}
  >{\raggedright\arraybackslash}p{(\linewidth - 12\tabcolsep) * \real{0.1429}}
  >{\raggedright\arraybackslash}p{(\linewidth - 12\tabcolsep) * \real{0.1429}}@{}}
\caption{Test accuracy for the 11 model configurations. MAE and RMSE are
in kWh; lower errors and higher \(R^2\) are better. Weather covariates
use KNMI observations as supplied in the original
benchmark.}\label{tbl-benchmark}\tabularnewline
\toprule\noalign{}
\begin{minipage}[b]{\linewidth}\raggedright
Model
\end{minipage} & \begin{minipage}[b]{\linewidth}\raggedright
Forecast-time inputs
\end{minipage} & \begin{minipage}[b]{\linewidth}\raggedright
MAE (kWh)
\end{minipage} & \begin{minipage}[b]{\linewidth}\raggedright
RMSE (kWh)
\end{minipage} & \begin{minipage}[b]{\linewidth}\raggedright
R²
\end{minipage} & \begin{minipage}[b]{\linewidth}\raggedright
MAPE (\%)
\end{minipage} & \begin{minipage}[b]{\linewidth}\raggedright
sMAPE (\%)
\end{minipage} \\
\midrule\noalign{}
\endfirsthead
\toprule\noalign{}
\begin{minipage}[b]{\linewidth}\raggedright
Model
\end{minipage} & \begin{minipage}[b]{\linewidth}\raggedright
Forecast-time inputs
\end{minipage} & \begin{minipage}[b]{\linewidth}\raggedright
MAE (kWh)
\end{minipage} & \begin{minipage}[b]{\linewidth}\raggedright
RMSE (kWh)
\end{minipage} & \begin{minipage}[b]{\linewidth}\raggedright
R²
\end{minipage} & \begin{minipage}[b]{\linewidth}\raggedright
MAPE (\%)
\end{minipage} & \begin{minipage}[b]{\linewidth}\raggedright
sMAPE (\%)
\end{minipage} \\
\midrule\noalign{}
\endhead
\bottomrule\noalign{}
\endlastfoot
TimesFM 3.0 with covariates & Calendar, lags and observed weather &
19.97 & 30.26 & 0.798 & 16.94 & 15.27 \\
TimesFM 3.0 target-only & Past consumption & 20.01 & 32.45 & 0.768 &
16.99 & 14.94 \\
TimesFM 2.5 target-only & Past consumption & 21.07 & 31.85 & 0.777 &
18.50 & 16.09 \\
XGBoost & Calendar, lags and observed weather & 22.45 & 31.11 & 0.787 &
18.98 & 16.86 \\
Prophet with regressors & Calendar, lags and observed weather & 24.91 &
34.22 & 0.742 & 21.39 & 19.73 \\
Ridge & Calendar, lags and observed weather & 27.18 & 37.87 & 0.684 &
22.93 & 20.43 \\
LSTM + ridge residual & Calendar, lags and observed weather & 30.07 &
43.53 & 0.583 & 24.85 & 22.36 \\
Prophet base & Calendar and holidays & 30.22 & 38.19 & 0.679 & 28.83 &
24.99 \\
LSTM base & Past consumption & 30.84 & 46.22 & 0.530 & 25.99 & 22.56 \\
Seasonal-naive 168 h & Past consumption & 35.27 & 53.26 & 0.376 & 29.29
& 24.80 \\
Seasonal-naive 24 h & Past consumption & 37.57 & 60.45 & 0.196 & 31.93 &
26.87 \\
\end{longtable}

TimesFM 3.0 with covariates has the lowest MAE (19.97 kWh), RMSE (30.26
kWh), and MAPE (16.94\%), with \(R^2=0.798\). Its MAE is almost
identical to the target-only version's 20.01 kWh, but its RMSE is lower
by 2.19 kWh. TimesFM 3.0 target-only has the lowest sMAPE (14.94\%).
XGBoost has MAE 22.45 kWh and RMSE 31.11 kWh; it therefore has higher
MAE but slightly lower RMSE than either target-only TimesFM version.

The weekly seasonal-naive forecast improves on the daily forecast, with
MAE 35.27 versus 37.57 kWh. Relative to the weekly reference, TimesFM
3.0 with covariates reduces MAE by 15.30 kWh, or 43.4\%. Prophet's
external regressors reduce MAE from 30.22 to 24.91 kWh. The LSTM
residual correction gives a smaller reduction, from 30.84 to 30.07 kWh.

\subsection{Differences Across Forecast
Days}\label{differences-across-forecast-days}

\begin{figure}

\centering{

\includegraphics[width=1\linewidth,height=\textheight,keepaspectratio,alt={Compact point-and-interval comparisons show uncertainty relative to XGBoost and between TimesFM configurations, with a labelled zero reference and direction of improvement.}]{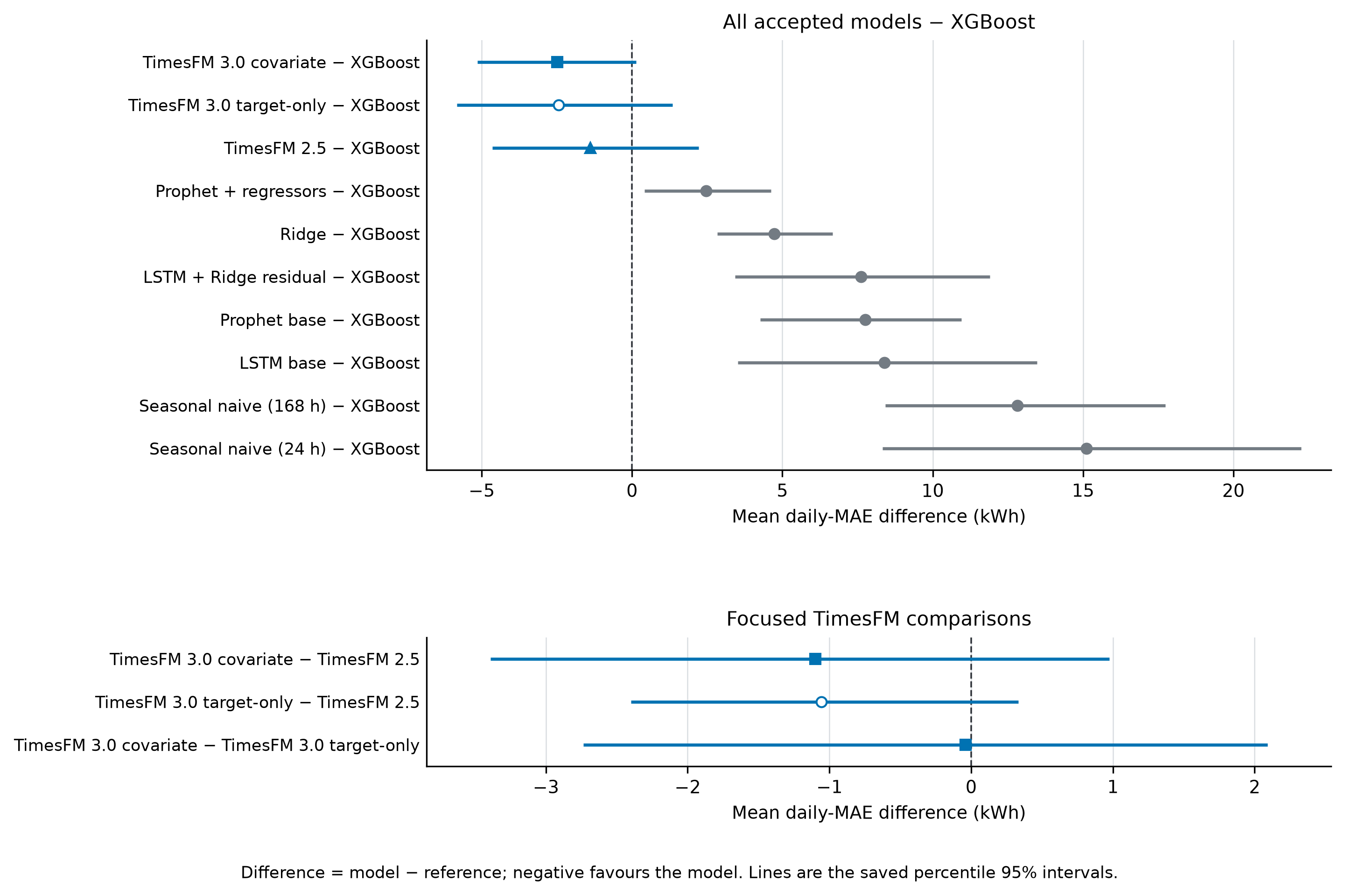}

}

\caption{\label{fig-paired-mae}Mean differences in daily MAE and their
95\% bootstrap intervals. The all-model comparison uses XGBoost as the
reference; the additional panel compares TimesFM configurations
directly. Differences are the named model minus its reference, so
negative values favour the named model. An interval crossing zero does
not establish a direction of improvement.}

\end{figure}

The paired comparisons in Figure~\ref{fig-paired-mae} show how the mean
daily-MAE differences vary when forecast days are resampled. The
numerical intervals are given in Table~\ref{tbl-paired-mae}.

\begin{longtable}[]{@{}
  >{\raggedright\arraybackslash}p{(\linewidth - 8\tabcolsep) * \real{0.2000}}
  >{\raggedright\arraybackslash}p{(\linewidth - 8\tabcolsep) * \real{0.2000}}
  >{\raggedright\arraybackslash}p{(\linewidth - 8\tabcolsep) * \real{0.2000}}
  >{\raggedright\arraybackslash}p{(\linewidth - 8\tabcolsep) * \real{0.2000}}
  >{\raggedright\arraybackslash}p{(\linewidth - 8\tabcolsep) * \real{0.2000}}@{}}
\caption{Mean daily-MAE differences from XGBoost and 95\% paired
bootstrap intervals, in kWh. Negative differences favour the listed
model; positive differences favour
XGBoost.}\label{tbl-paired-mae}\tabularnewline
\toprule\noalign{}
\begin{minipage}[b]{\linewidth}\raggedright
Model
\end{minipage} & \begin{minipage}[b]{\linewidth}\raggedright
Reference
\end{minipage} & \begin{minipage}[b]{\linewidth}\raggedright
Mean daily-MAE difference (kWh)
\end{minipage} & \begin{minipage}[b]{\linewidth}\raggedright
95\% percentile CI (kWh)
\end{minipage} & \begin{minipage}[b]{\linewidth}\raggedright
Paired days
\end{minipage} \\
\midrule\noalign{}
\endfirsthead
\toprule\noalign{}
\begin{minipage}[b]{\linewidth}\raggedright
Model
\end{minipage} & \begin{minipage}[b]{\linewidth}\raggedright
Reference
\end{minipage} & \begin{minipage}[b]{\linewidth}\raggedright
Mean daily-MAE difference (kWh)
\end{minipage} & \begin{minipage}[b]{\linewidth}\raggedright
95\% percentile CI (kWh)
\end{minipage} & \begin{minipage}[b]{\linewidth}\raggedright
Paired days
\end{minipage} \\
\midrule\noalign{}
\endhead
\bottomrule\noalign{}
\endlastfoot
Seasonal-naive 24 h & XGBoost & 15.12 & {[}8.338, 22.252{]} & 57 \\
Seasonal-naive 168 h & XGBoost & 12.82 & {[}8.421, 17.742{]} & 57 \\
Ridge & XGBoost & 4.73 & {[}2.841, 6.671{]} & 57 \\
Prophet base & XGBoost & 7.76 & {[}4.260, 10.958{]} & 57 \\
Prophet with regressors & XGBoost & 2.46 & {[}0.422, 4.622{]} & 57 \\
TimesFM 2.5 target-only & XGBoost & -1.38 & {[}-4.636, 2.218{]} & 57 \\
TimesFM 3.0 target-only & XGBoost & -2.44 & {[}-5.825, 1.342{]} & 57 \\
TimesFM 3.0 with covariates & XGBoost & -2.48 & {[}-5.140, 0.132{]} &
57 \\
LSTM base & XGBoost & 8.39 & {[}3.526, 13.461{]} & 57 \\
LSTM + ridge residual & XGBoost & 7.62 & {[}3.427, 11.903{]} & 57 \\
\end{longtable}

TimesFM 3.0 with covariates has a mean daily-MAE difference of -2.48 kWh
from XGBoost, but its interval spans -5.14 to 0.13 kWh. The target-only
TimesFM 3.0 and TimesFM 2.5 comparisons also include zero. Direct
comparisons between TimesFM 3.0 and 2.5 are similarly inconclusive. The
native-covariate minus target-only TimesFM 3.0 difference is only -0.04
kWh, with an interval from -2.735 to 2.090 kWh. Thus, the lowest pooled
point estimate does not establish a clear daily-MAE advantage among
these configurations over the 57 days.

The intervals comparing both seasonal-naive forecasts, ridge, both
Prophet configurations, and both LSTM configurations with XGBoost are
wholly positive. Within this test, these comparisons provide clearer
evidence of lower mean daily MAE for XGBoost. They remain comparisons of
the particular fitted configurations and covariates used here.

\subsection{Workdays, Weekends, and
Holidays}\label{workdays-weekends-and-holidays}

\begin{figure}

\centering{

\includegraphics[width=1\linewidth,height=\textheight,keepaspectratio,alt={Six aligned dot panels compare MAE and RMSE across workdays, weekends, and holidays, making the holiday benefit and routine-day cost of TimesFM covariates visible.}]{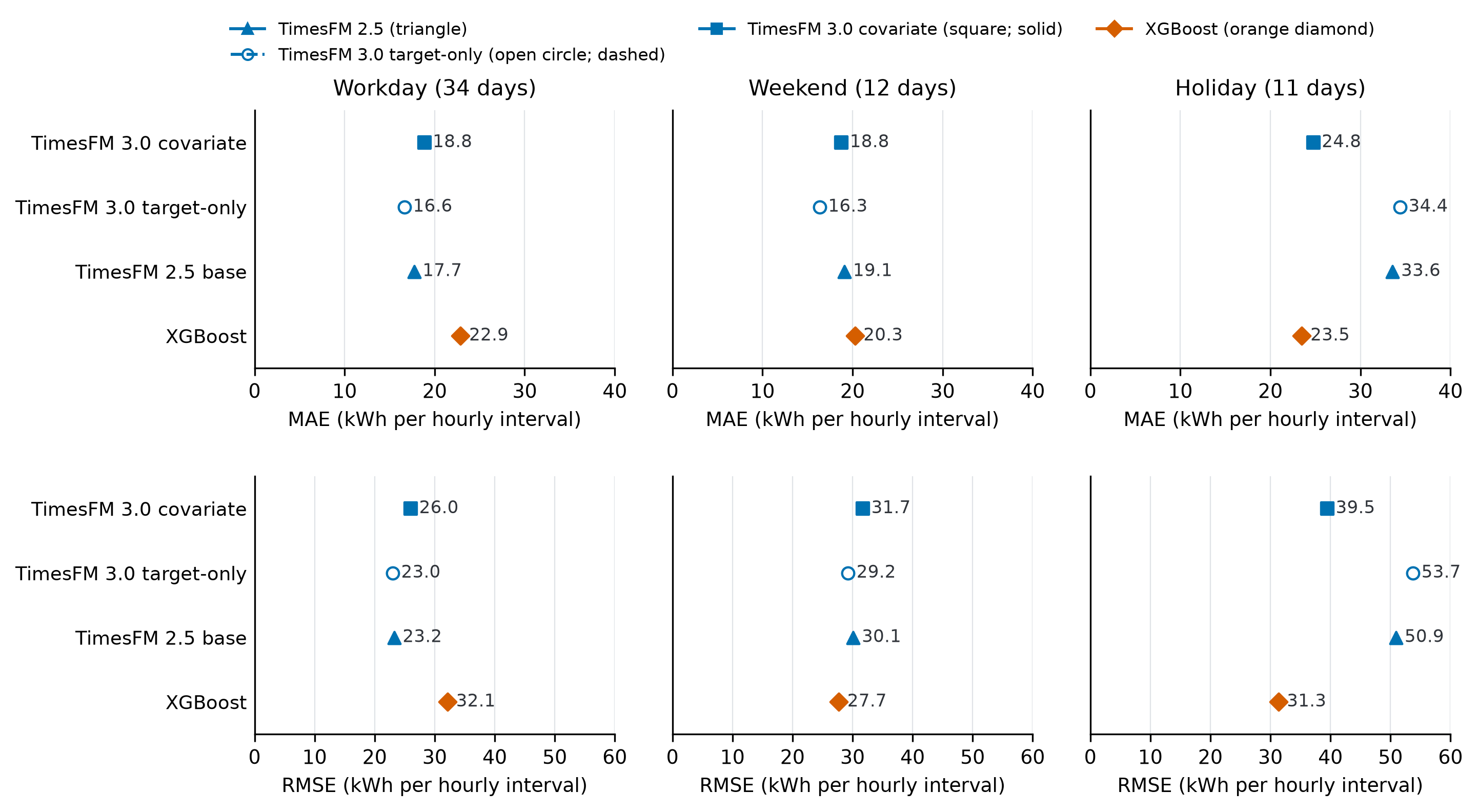}

}

\caption{\label{fig-regime-errors}MAE and RMSE for TimesFM 2.5, TimesFM
3.0 with and without covariates, and XGBoost, evaluated separately on 34
workdays, 12 weekend days, and 11 holidays. Model settings are identical
across groups. Each metric uses a common scale across the three day
types; subgroup differences are descriptive.}

\end{figure}

The test contains 34 workdays (816 hours) and 23 non-workdays (552
hours). Non-workdays comprise 12 weekend days and 11 holidays.
Table~\ref{tbl-regime-errors} reports the complete model comparison.
Figure~\ref{fig-regime-errors} separates workdays, weekends, and
holidays for the three TimesFM configurations and XGBoost, showing how
their relative performance changes with day type and error measure. The
corresponding figure for all 11 configurations is provided in Appendix
Section~\ref{sec-full-daytype-comparison}.

\begin{longtable}[]{@{}
  >{\raggedright\arraybackslash}p{(\linewidth - 8\tabcolsep) * \real{0.2000}}
  >{\raggedright\arraybackslash}p{(\linewidth - 8\tabcolsep) * \real{0.2000}}
  >{\raggedright\arraybackslash}p{(\linewidth - 8\tabcolsep) * \real{0.2000}}
  >{\raggedright\arraybackslash}p{(\linewidth - 8\tabcolsep) * \real{0.2000}}
  >{\raggedright\arraybackslash}p{(\linewidth - 8\tabcolsep) * \real{0.2000}}@{}}
\caption{Pooled test errors by day type, in kWh. Subgroup rankings
describe the observed errors and have not been tested for statistical
significance.}\label{tbl-regime-errors}\tabularnewline
\toprule\noalign{}
\begin{minipage}[b]{\linewidth}\raggedright
Model
\end{minipage} & \begin{minipage}[b]{\linewidth}\raggedright
Workday MAE (kWh)
\end{minipage} & \begin{minipage}[b]{\linewidth}\raggedright
Workday RMSE (kWh)
\end{minipage} & \begin{minipage}[b]{\linewidth}\raggedright
Non-workday MAE (kWh)
\end{minipage} & \begin{minipage}[b]{\linewidth}\raggedright
Non-workday RMSE (kWh)
\end{minipage} \\
\midrule\noalign{}
\endfirsthead
\toprule\noalign{}
\begin{minipage}[b]{\linewidth}\raggedright
Model
\end{minipage} & \begin{minipage}[b]{\linewidth}\raggedright
Workday MAE (kWh)
\end{minipage} & \begin{minipage}[b]{\linewidth}\raggedright
Workday RMSE (kWh)
\end{minipage} & \begin{minipage}[b]{\linewidth}\raggedright
Non-workday MAE (kWh)
\end{minipage} & \begin{minipage}[b]{\linewidth}\raggedright
Non-workday RMSE (kWh)
\end{minipage} \\
\midrule\noalign{}
\endhead
\bottomrule\noalign{}
\endlastfoot
TimesFM 3.0 with covariates & 18.85 & 26.00 & 21.63 & 35.63 \\
TimesFM 3.0 target-only & 16.65 & 23.00 & 24.98 & 42.75 \\
TimesFM 2.5 target-only & 17.71 & 23.25 & 26.03 & 41.41 \\
XGBoost & 22.88 & 32.15 & 21.81 & 29.50 \\
Prophet with regressors & 23.18 & 31.56 & 27.47 & 37.80 \\
Ridge & 27.41 & 37.88 & 26.84 & 37.85 \\
LSTM + ridge residual & 30.26 & 44.31 & 29.80 & 42.35 \\
Prophet base & 25.88 & 31.97 & 36.62 & 45.87 \\
LSTM base & 30.26 & 46.33 & 31.70 & 46.07 \\
Seasonal-naive 168 h & 32.29 & 47.59 & 39.68 & 60.67 \\
Seasonal-naive 24 h & 36.27 & 57.75 & 39.50 & 64.25 \\
\end{longtable}

TimesFM 3.0 target-only has the lowest workday MAE and RMSE, 16.65 and
23.00 kWh. On non-workdays, TimesFM 3.0 with covariates has slightly
lower MAE than XGBoost (21.63 versus 21.81 kWh), but higher RMSE (35.63
versus 29.50 kWh).

Within TimesFM 3.0, adding covariates reduces holiday RMSE from 53.74 to
39.47 kWh. In contrast, workday RMSE increases from 23.00 to 26.00 kWh
and weekend RMSE from 29.25 to 31.71 kWh.

\begin{figure}

\centering{

\includegraphics[width=1\linewidth,height=\textheight,keepaspectratio,alt={Three hourly forecast examples show observations alongside TimesFM 3.0 with and without covariates and XGBoost for the first workday, weekend, and holiday test windows.}]{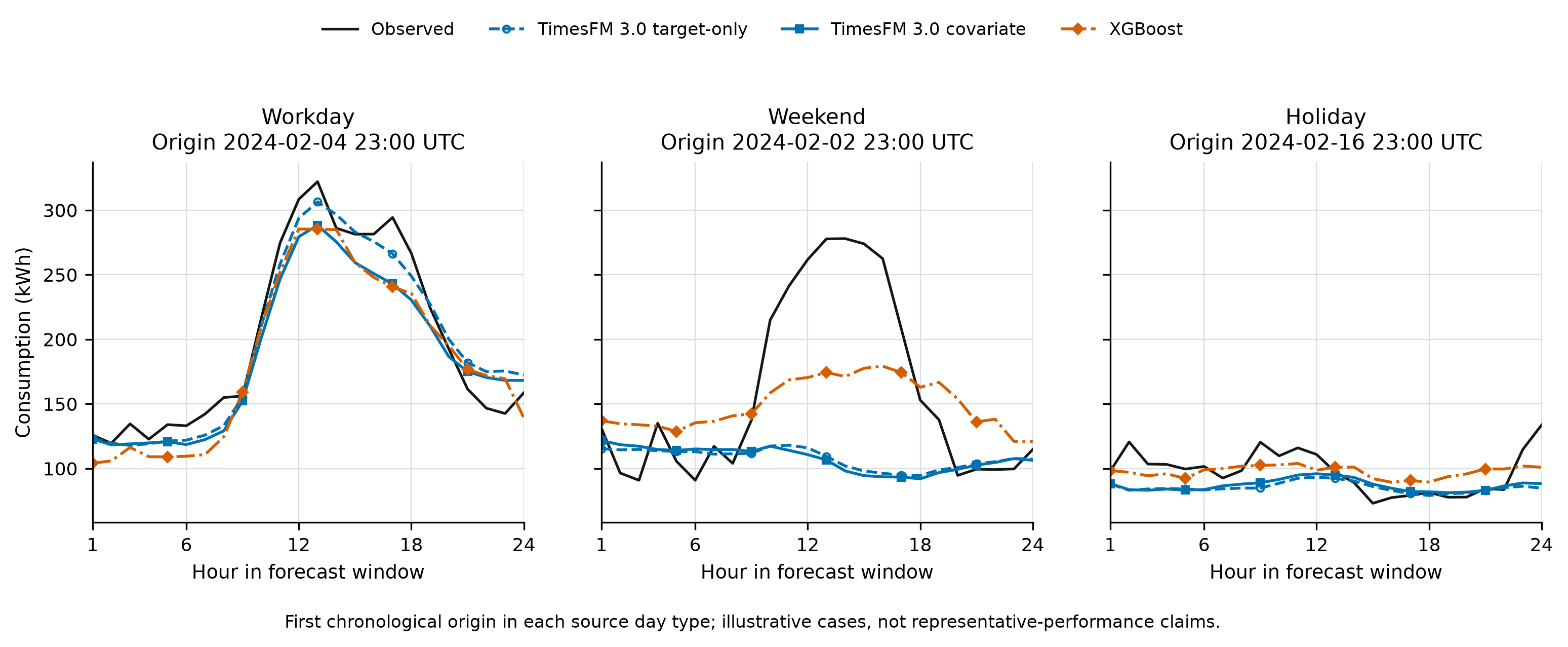}

}

\caption{\label{fig-forecast-examples}Observed electricity consumption
and frozen TimesFM 3.0 and XGBoost forecasts for the first chronological
test forecast origin in each day-type group. The origin labels are in
UTC, and each panel contains the corresponding 24 hourly targets on a
common consumption scale. Black identifies observations; colour and line
style identify the forecasts. The selection rule does not use forecast
errors.}

\end{figure}

Figure~\ref{fig-forecast-examples} complements these aggregate errors
with the hourly profiles from the first test forecast in each day-type
group. The observed and forecast curves show differences in predicted
consumption level and timing within each window. The complete group
scores above provide the basis for comparing model performance.

\subsection{Extended Historical
Evaluation}\label{sec-historical-evaluation}

The extended historical evaluation covers 566 forecast origins and
13,584 hourly targets, from 11 September 2022 at 23:00 UTC through 30
March 2024 at 22:00 UTC. The first seven days supply the initial
context. TimesFM 3.0 with covariates gives MAE 16.60 kWh and RMSE 26.74
kWh, compared with 29.16 and 49.03 kWh for the weekly seasonal-naive
forecast (Table~\ref{tbl-replay}). The 95th percentile of absolute error
is 48.52 kWh for TimesFM and 115.38 kWh for the seasonal-naive forecast.

\begin{longtable}[]{@{}llllll@{}}
\caption{Extended historical evaluation on 13,584 common target hours.
Error values are in kWh. The TimesFM settings were selected using later
validation data; these scores are
exploratory.}\label{tbl-replay}\tabularnewline
\toprule\noalign{}
Scenario & n & MAE & RMSE & R² & p95 absolute error \\
\midrule\noalign{}
\endfirsthead
\toprule\noalign{}
Scenario & n & MAE & RMSE & R² & p95 absolute error \\
\midrule\noalign{}
\endhead
\bottomrule\noalign{}
\endlastfoot
TimesFM 3.0 with covariates & 13584 & 16.6 & 26.74 & 0.879 & 48.52 \\
Seasonal-naive 168 h & 13584 & 29.16 & 49.03 & 0.593 & 115.38 \\
\end{longtable}

\begin{figure}

\centering{

\includegraphics[width=1\linewidth,height=\textheight,keepaspectratio,alt={Vertically aligned monthly MAE and RMSE curves show seasonal changes for TimesFM and the weekly seasonal-naive comparator, with shared date labels.}]{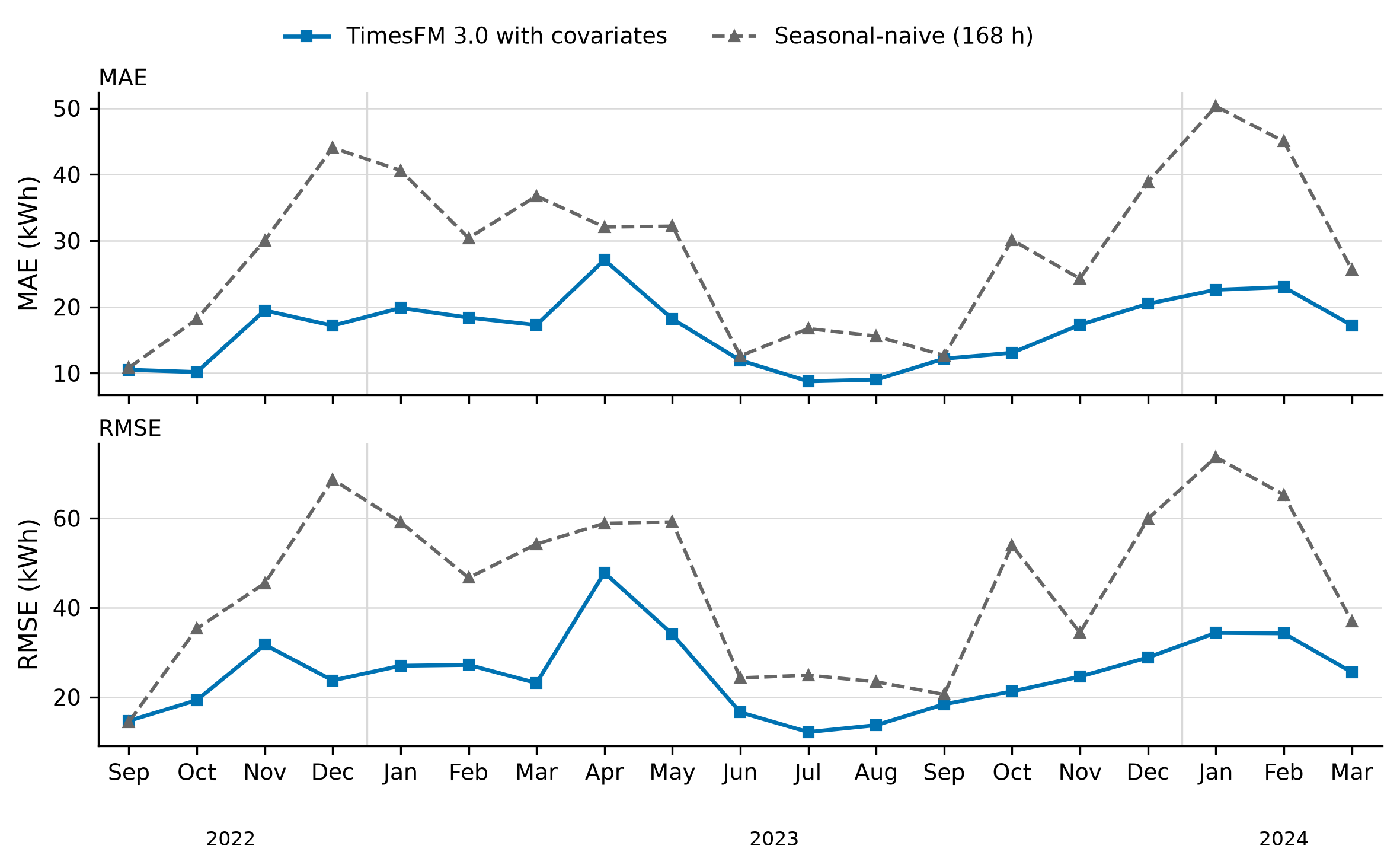}

}

\caption{\label{fig-replay-monthly}Monthly pooled MAE and RMSE for
TimesFM 3.0 with covariates and the weekly seasonal-naive forecast
during the exploratory historical evaluation. The aligned panels share a
chronological month axis, and the two models use different line styles
and markers. Months refer to UTC target timestamps; September 2022 and
March 2024 are partial months.}

\end{figure}

Seasonal errors vary substantially. TimesFM MAE is 20.25 kWh in winter,
19.92 kWh in spring, 9.89 kWh in summer, and 13.98 kWh in autumn. The
chronological monthly comparison in Figure~\ref{fig-replay-monthly}
shows this variation alongside the seasonal-naive reference. These
groups have unequal coverage, and the available TimesFM context grows
during the first year.

The historical evaluation contains 330 workdays, 121 weekend days, and
115 holiday days. TimesFM MAE is 16.86, 14.65, and 17.89 kWh,
respectively. Holiday RMSE remains highest at 34.97 kWh, compared with
23.81 kWh on workdays and 25.23 kWh on weekends. The corresponding
pooled \(R^2\) is 0.910 on workdays but -0.086 on holidays. Appendix
Section~\ref{sec-supplementary-breakdowns} provides the counts, monthly
context ranges, and season/day-type tables.

\subsection{Weather-Source
Sensitivity}\label{weather-source-sensitivity}

\begin{figure}

\centering{

\includegraphics[width=1\linewidth,height=\textheight,keepaspectratio,alt={Three horizontal point-and-interval comparisons show an ERA5-minus-KNMI difference near zero and small positive historical-IFS differences.}]{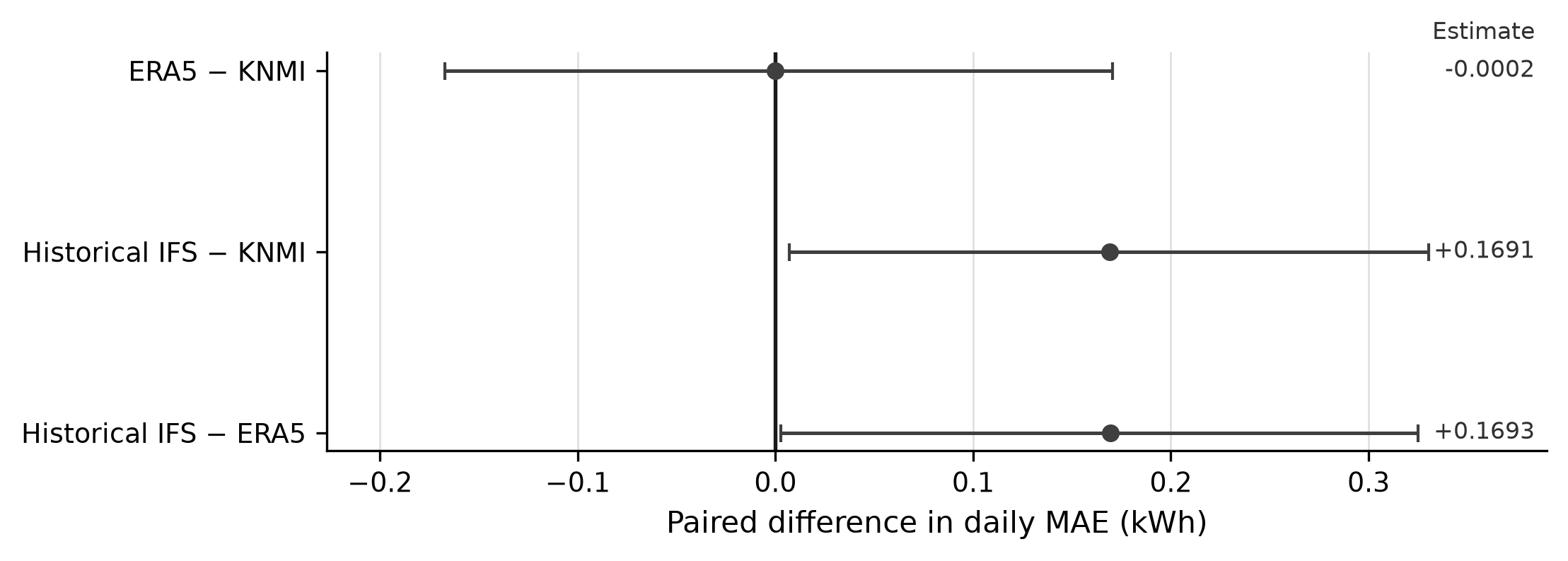}

}

\caption{\label{fig-weather-sensitivity}Paired differences in daily MAE
between weather sources, with 95\% day-block bootstrap intervals. The
zero line denotes no difference; a positive value indicates larger error
for the first source named. All comparisons use the same 57 forecast
windows and fixed TimesFM 3.0 settings. Absolute error values are
reported in Table~\ref{tbl-weather-accuracy}.}

\end{figure}

All three weather sources produce complete forecasts for the same 57
windows and 1,368 target hours. Table~\ref{tbl-weather-accuracy} reports
their absolute errors, while Figure~\ref{fig-weather-sensitivity}
isolates the paired differences between sources. MAE is 19.968 kWh with
KNMI observations, 19.968 kWh with ERA5, and 20.137 kWh with historical
IFS. Corresponding RMSE values are 30.288, 30.221, and 30.409 kWh. ERA5
has the lowest point estimates, although its MAE differs from KNMI by
less than 0.001 kWh.

\begin{longtable}[]{@{}
  >{\raggedright\arraybackslash}p{(\linewidth - 12\tabcolsep) * \real{0.1429}}
  >{\raggedright\arraybackslash}p{(\linewidth - 12\tabcolsep) * \real{0.1429}}
  >{\raggedright\arraybackslash}p{(\linewidth - 12\tabcolsep) * \real{0.1429}}
  >{\raggedright\arraybackslash}p{(\linewidth - 12\tabcolsep) * \real{0.1429}}
  >{\raggedright\arraybackslash}p{(\linewidth - 12\tabcolsep) * \real{0.1429}}
  >{\raggedright\arraybackslash}p{(\linewidth - 12\tabcolsep) * \real{0.1429}}
  >{\raggedright\arraybackslash}p{(\linewidth - 12\tabcolsep) * \real{0.1429}}@{}}
\caption{Fixed TimesFM 3.0 weather-source comparison using the same UTC
hourly intervals. Context, other covariates, and model settings are held
constant. This KNMI result uses the supplementary interval assignment
and is distinct from the original benchmark
row.}\label{tbl-weather-accuracy}\tabularnewline
\toprule\noalign{}
\begin{minipage}[b]{\linewidth}\raggedright
Source
\end{minipage} & \begin{minipage}[b]{\linewidth}\raggedright
Target hours
\end{minipage} & \begin{minipage}[b]{\linewidth}\raggedright
MAE (kWh)
\end{minipage} & \begin{minipage}[b]{\linewidth}\raggedright
RMSE (kWh)
\end{minipage} & \begin{minipage}[b]{\linewidth}\raggedright
R²
\end{minipage} & \begin{minipage}[b]{\linewidth}\raggedright
MAPE (\%)
\end{minipage} & \begin{minipage}[b]{\linewidth}\raggedright
sMAPE (\%)
\end{minipage} \\
\midrule\noalign{}
\endfirsthead
\toprule\noalign{}
\begin{minipage}[b]{\linewidth}\raggedright
Source
\end{minipage} & \begin{minipage}[b]{\linewidth}\raggedright
Target hours
\end{minipage} & \begin{minipage}[b]{\linewidth}\raggedright
MAE (kWh)
\end{minipage} & \begin{minipage}[b]{\linewidth}\raggedright
RMSE (kWh)
\end{minipage} & \begin{minipage}[b]{\linewidth}\raggedright
R²
\end{minipage} & \begin{minipage}[b]{\linewidth}\raggedright
MAPE (\%)
\end{minipage} & \begin{minipage}[b]{\linewidth}\raggedright
sMAPE (\%)
\end{minipage} \\
\midrule\noalign{}
\endhead
\bottomrule\noalign{}
\endlastfoot
KNMI station 344 & 1368 & 19.9682 & 30.2878 & 0.7981 & 16.9134 &
15.2548 \\
ERA5 reanalysis & 1368 & 19.968 & 30.2208 & 0.799 & 16.9045 & 15.2468 \\
Historical IFS & 1368 & 20.1373 & 30.4094 & 0.7965 & 17.1393 & 15.389 \\
\end{longtable}

The ERA5-minus-KNMI mean daily-MAE difference is effectively zero, with
a 95\% interval of -0.167 to 0.171 kWh. Historical IFS has a difference
of 0.169 kWh from KNMI, with an interval of 0.007 to 0.331 kWh, and
0.169 kWh from ERA5, with an interval of 0.003 to 0.325 kWh. Thus the
unadjusted intervals indicate a small increase for historical IFS,
approximately 0.85\% of KNMI MAE. Their lower bounds lie close to zero,
and the analysis does not adjust for multiple comparisons or dependence
between successive days.

The day-type pattern is consistent across sources: holiday MAE is
approximately 24.74--25.29 kWh, compared with 18.84--18.97 kWh on
workdays. Appendix Section~\ref{sec-supplementary-breakdowns} reports
the day-type figure, monthly/seasonal errors, and numerical paired
intervals.

\subsection{Fitting and Forecasting
Time}\label{fitting-and-forecasting-time}

\begin{longtable}[]{@{}
  >{\raggedright\arraybackslash}p{(\linewidth - 8\tabcolsep) * \real{0.2000}}
  >{\raggedright\arraybackslash}p{(\linewidth - 8\tabcolsep) * \real{0.2000}}
  >{\raggedright\arraybackslash}p{(\linewidth - 8\tabcolsep) * \real{0.2000}}
  >{\raggedright\arraybackslash}p{(\linewidth - 8\tabcolsep) * \real{0.2000}}
  >{\raggedright\arraybackslash}p{(\linewidth - 8\tabcolsep) * \real{0.2000}}@{}}
\caption{Measured time and memory during test forecasting. Times are
seconds per 24-hour forecast origin, summarised by median and 95th
percentile. The LSTM ridge-correction row is additional to the LSTM base
cost.}\label{tbl-resources}\tabularnewline
\toprule\noalign{}
\begin{minipage}[b]{\linewidth}\raggedright
Model
\end{minipage} & \begin{minipage}[b]{\linewidth}\raggedright
Fit/update median / p95 (s)
\end{minipage} & \begin{minipage}[b]{\linewidth}\raggedright
Warm 24-h inference median / p95 (s)
\end{minipage} & \begin{minipage}[b]{\linewidth}\raggedright
Peak RSS (MiB)
\end{minipage} & \begin{minipage}[b]{\linewidth}\raggedright
Peak GPU allocated (MiB)
\end{minipage} \\
\midrule\noalign{}
\endfirsthead
\toprule\noalign{}
\begin{minipage}[b]{\linewidth}\raggedright
Model
\end{minipage} & \begin{minipage}[b]{\linewidth}\raggedright
Fit/update median / p95 (s)
\end{minipage} & \begin{minipage}[b]{\linewidth}\raggedright
Warm 24-h inference median / p95 (s)
\end{minipage} & \begin{minipage}[b]{\linewidth}\raggedright
Peak RSS (MiB)
\end{minipage} & \begin{minipage}[b]{\linewidth}\raggedright
Peak GPU allocated (MiB)
\end{minipage} \\
\midrule\noalign{}
\endhead
\bottomrule\noalign{}
\endlastfoot
TimesFM 3.0 with covariates & 0.000000 / 0.000000 & 0.9215 / 0.9680 &
4402.73 & 2470.39 \\
TimesFM 3.0 target-only & 0.000000 / 0.000000 & 0.4047 / 0.4566 &
4402.73 & 1308.73 \\
TimesFM 2.5 target-only & 0.000000 / 0.000000 & 0.1197 / 0.1394 &
2435.05 & 898.29 \\
XGBoost & 0.8418 / 3.43 & 0.001232 / 0.004213 & 378.00 & not
applicable \\
Prophet with regressors & 4.19 / 5.97 & 0.3201 / 0.3582 & 254.04 & not
applicable \\
Ridge & 0.0302 / 0.0359 & 0.002373 / 0.003514 & 191.07 & not
applicable \\
LSTM + ridge (correction only) & 0.007708 / 0.009089 & 0.001936 /
0.002448 & 3443.93 & not recorded \\
Prophet base & 2.09 / 2.84 & 0.2969 / 0.3550 & 235.48 & not
applicable \\
LSTM base & 99.05 / 204.05 & 0.1497 / 0.1789 & 3443.93 & not recorded \\
Seasonal-naive 168 h & 0.000000 / 0.000000 & 0.000080 / 0.000104 &
188.10 & not applicable \\
Seasonal-naive 24 h & 0.000000 / 0.000000 & 0.000081 / 0.000135 & 184.08
& not applicable \\
\end{longtable}

TimesFM's model weights were trained externally and reused for this
building. In our execution, the median forecast time was 0.12 s for
TimesFM 2.5, 0.41 s for target-only TimesFM 3.0, and 0.92 s for TimesFM
3.0 with covariates. Loading TimesFM 3.0 took approximately two seconds
before forecasting. These measurements exclude its original pretraining.

Ridge, XGBoost, Prophet, and the LSTM required building-specific fitting
at each origin under the evaluated policy. Median fitting times were
approximately 0.03 s for ridge, 0.84 s for XGBoost, 2.10 s for Prophet
base, 4.19 s for Prophet with regressors, and 99.05 s for the LSTM. The
ridge correction added approximately 0.008 s to the LSTM fit.

\section{Discussion}\label{discussion}

\subsection{Accuracy in Relation to Previous
Studies}\label{accuracy-in-relation-to-previous-studies}

TimesFM provides a strong starting point for building electricity
forecasting without local neural-network training. TimesFM 3.0 with
covariates ranks first in pooled MAE and RMSE, while the target-only 3.0
and 2.5 models use only past electricity consumption and also achieve
lower pooled MAE than all locally fitted configurations. XGBoost is the
closest locally fitted alternative, with pooled RMSE only 0.85 kWh above
TimesFM 3.0 with covariates. These results demonstrate that pretrained
forecasting can deliver leading accuracy without fitting model weights
to the building. The paired daily-MAE intervals leave the ranking of
TimesFM and XGBoost uncertain, while the LSTM's longer local fitting
brings no accuracy advantage under the evaluated policy.

The preceding THUAS evaluation found the lowest workday MAE with Prophet
and the lowest MAPE and mean squared error with XGBoost (Stienstra et
al., 2026). The present comparison adds pretrained forecasting and
evaluates the selected configurations over 57 days including closures.
TimesFM's strong performance extends the evidence for zero-shot transfer
reported by Das et al. (2024) and the electrical-demand comparison by
Llanes-Guilarte et al. (2026) to this university building. XGBoost
retains an advantage when larger non-workday errors are the priority:
its non-workday RMSE is 6.13 kWh lower than that of TimesFM 3.0 with
covariates, despite their MAEs differing by only 0.18 kWh. The preferred
error profile therefore depends on the operating period and the
consequences of larger misses.

TimesFM 2.5 already provides a strong target-only forecast, with MAE
21.07 kWh. Version 3.0 reduces target-only MAE to 20.01 kWh, although
its RMSE is slightly higher, at 32.45 versus 31.85 kWh. The
native-covariate 3.0 configuration gives the lowest pooled values for
both measures. The extended historical evaluation and weather-source
comparison further assess this selected configuration across seasons and
alternative inputs. Together, these results support using pretrained
TimesFM as an initial forecasting baseline alongside seasonal-naive
references, before investing in building-specific model training.

\subsection{What the Covariates Add}\label{what-the-covariates-add}

The benefit of covariates depends on how effectively a model already
uses consumption history. External regressors reduce Prophet's pooled
MAE by 5.31 kWh, ridge residual correction reduces LSTM MAE by 0.77 kWh,
and native covariates reduce TimesFM 3.0 MAE by only 0.04 kWh. The
TimesFM result indicates two complementary strengths: the consumption
series contains recurring daily and weekly operating patterns, and the
pretrained model can use those patterns effectively with its weights
held fixed. Together, these provide a plausible explanation for the
small overall gain from explicitly supplying calendar, weather, and
historical-consumption covariates.

For TimesFM, the main benefit appears during holidays: adding the
covariate set reduces MAE from 34.40 to 24.76 kWh and RMSE from 53.74 to
39.47 kWh. The 14.27 kWh reduction in holiday RMSE accompanies smaller
increases of 3.00 kWh on workdays and 2.46 kWh on weekends, yielding a
lower pooled RMSE. These gains were obtained with the complete calendar,
weather, and historical-consumption covariate set. The results favour
this configuration when reducing holiday errors is a priority, despite
its smaller loss of accuracy during routine operation.

The practical choice is therefore straightforward: target-only TimesFM
provides strong forecasts from consumption history alone, while its
covariate version improves closure-period forecasts at the cost of
additional input preparation. This distinction separates the value of a
simple initial forecast from the value of refining performance during
specific operating periods.

\subsection{Performance Across the
Year}\label{performance-across-the-year}

The extended historical evaluation uses a practical advantage of
pretrained forecasting: TimesFM can forecast after a seven-day context
warm-up without fitting new network weights. This permits examination of
periods otherwise reserved for the benchmark's 365-day local fitting
window. Its MAE varies from 9.89 kWh in summer to 20.25 kWh in winter,
showing the value of examining the full operating year when assessing
forecasting difficulty. The comparison with weekly seasonal-naive
forecasts adds a reference for each period. Seasonal differences should
be interpreted alongside the unequal coverage and growing context during
the first year; the analysis extends the assessment of TimesFM rather
than the ranking of all 11 configurations.

The historical holiday results also clarify the role of accuracy
metrics. Holiday MAE is close to workday MAE, but holiday \(R^2\) is
slightly negative because squared forecast errors exceed the variation
around the holiday group's mean. Lower consumption variability makes
this reference harder to improve upon, while occasional large errors
increase the squared-error total. MAE and RMSE remain directly
interpretable in hourly kWh, whereas \(R^2\) describes performance
relative to variation within the group. Reporting them together makes
the closure-period errors easier to assess.

\subsection{Weather Inputs and Practical Model
Choice}\label{weather-inputs-and-practical-model-choice}

The weather experiment addresses a practical concern about the
benchmark's use of station observations: would a different weather
product substantially change the TimesFM forecasts? For the tested
configuration, replacing KNMI with ERA5 changes MAE by less than 0.001
kWh, and replacing it with historical IFS increases MAE by approximately
0.17 kWh, or 0.85\%. The IFS increase is small in magnitude even though
its unadjusted paired interval excludes zero. This provides evidence of
limited sensitivity to the three tested weather products, rather than
evidence of identical predictions or exactly equal accuracy.

This finding strengthens the case for investigating pretrained
forecasting at THUAS: the covariate configuration's accuracy is not
strongly dependent on choosing one of these weather sources. The small
source differences also persist within day types, including holidays,
where covariates provide the clearest improvement over target-only
TimesFM. However, weather-source substitution and covariate inclusion
answer different questions. All three source variants contain weather,
and the other 23 covariates remain unchanged. The experiment does not
show that weather is unimportant, that reanalysis is necessarily more
accurate at the building, or that another forecasting model would have
the same sensitivity.

Our comparison replaces both historical and future weather together and
does not reconstruct issue-time-specific inputs. Target-only TimesFM
avoids the weather-data requirement entirely, providing a simpler input
pipeline; the covariate configuration offers the holiday-error
improvement discussed above. Any deployment must also respect the
checkpoint's restrictions noted below.

\subsection{Local Modelling Effort}\label{local-modelling-effort}

Pretrained forecasting replaces local weight fitting with
historical-context preparation and inference. In this benchmark, that
change does not entail an observed pooled-accuracy penalty relative to
XGBoost: TimesFM 3.0 with covariates has lower point estimates, with
uncertainty that prevents a clear daily-MAE ranking. The target-only
version further reduces the number of data sources needed. Nevertheless,
both TimesFM configurations underwent validation of context and
inference settings; zero-shot forecasting is not equivalent to an
untuned or zero-history application.

XGBoost offers a different practical advantage. After a median fit of
0.84 s, its measured 24-hour inference took approximately 0.0012 s,
compared with 0.92 s for TimesFM 3.0 with covariates in the recorded
CPU/GPU environments. Both inference times are short relative to a daily
forecasting cycle, making their difference less consequential for one
building than for a larger forecasting service. Under the evaluated
policy, XGBoost is refitted each day as new observations arrive,
requiring an ongoing fitting procedure and maintenance of its covariate
inputs. TimesFM incorporates new observations by advancing its context
without refitting network weights. Both approaches still require checks
on input quality and forecast performance. The LSTM's approximately 99 s
median fit is appreciably longer, without lower errors in this
benchmark.

\section{Limitations}\label{limitations}

The study is designed to support a site-specific model comparison, not a
universal ranking of forecasting methods. Using the same target hours,
chronological model selection, seasonal-naive references, and paired
daily comparisons makes the 57-day test informative for this building
despite its limited duration. The test is weighted toward late winter
and early spring, however, and contains only 11 holiday days. Its
bootstrap intervals are not adjusted for multiple comparisons and do not
model dependence between successive days. The extended historical
evaluation adds seasonal and closure-period evidence, but uses shorter
contexts at early origins and settings selected from later validation
data; summer is represented only once. It therefore complements rather
than replaces an independent all-season test.

Rotterdam Airport observations were used because they are the regional
weather source in the supplied dataset. Comparing them with two gridded
products directly tests whether dependence on that source substantially
changes the selected TimesFM results; the small measured differences
reduce that concern for the evaluated period. They do not establish
exact weather conditions at the building or test other stations. The
products differ in temporal sampling, and the supplementary comparison
uses a consistent endpoint assignment while the original benchmark
retains the supplied weather join. Independent documentation of the
original meter interval-label convention was unavailable (Appendix
Section~\ref{sec-time-weather-alignment}). Historical IFS identifies a
forecast product, but its downloaded values lack the issue times needed
to reconstruct a fully operational forecast sequence.

Treating opening hours and examination weeks as known one day ahead is a
reasonable planning assumption for scheduled university activities. We
exclude realised attendance and cumulative room bookings rather than
treating tomorrow's observed activity as an available input.
Nevertheless, the historical publication times of the supplied schedules
were not retained, and calendar labels do not measure actual attendance.
The results therefore apply to the stated schedule-availability
assumption.

The resource measurements distinguish recurring fitting from
forecasting, which is sufficient to compare those tasks within the
recorded executions. They do not establish total resource efficiency:
LSTM GPU peaks and some model-storage measurements are missing, runs use
different CPU/GPU settings, and development and pretraining costs are
excluded. Finally, the evaluated TimesFM 3.0 weights are released for
non-commercial, non-production research use (Google Research, 2026). The
results can motivate operational assessment, but do not establish
permission to deploy this checkpoint in production.

\section{Conclusion}\label{conclusion}

Pretrained TimesFM and locally fitted XGBoost provide the strongest
alternatives among the 11 configurations evaluated for this university
building. TimesFM 3.0 with covariates achieved the lowest pooled MAE and
RMSE, 19.97 and 30.26 kWh per hourly target, after chronological
validation and evaluation on 57 test days. Its MAE was 43.4\% lower than
the weekly seasonal-naive reference. XGBoost remained close in RMSE at
31.11 kWh, with rapid inference after fitting and no clear separation
from TimesFM in the paired daily-MAE comparisons.

Workday and non-workday errors lead to different choices: target-only
TimesFM 3.0 had the lowest workday errors, whereas XGBoost had the
lowest non-workday RMSE. Adding covariates to TimesFM substantially
reduced holiday errors but provided almost no pooled MAE improvement.
The exploratory 566-day historical evaluation extended the analysis
beyond the winter test and revealed substantial seasonal variation
without additional network training. Replacing KNMI observations with
ERA5 or historical IFS changed MAE by at most 0.17 kWh, showing limited
weather-source sensitivity within the selected covariate configuration.

Target-only TimesFM therefore offers a strong initial forecast for this
building from past electricity consumption alone, without local weight
training or external covariate preparation. Adding covariates
substantially improves holiday forecasts, with smaller increases in
workday and weekend errors. XGBoost remains an accurate, fast
alternative when local fitting and ongoing covariate maintenance are
acceptable. The practical starting point is a pretrained target-only
forecast, followed by comparison with a locally fitted model and the
addition of covariates where closure-period accuracy justifies the extra
inputs.

\section{Future Work}\label{future-work}

Future work should evaluate the selected approaches on independent
buildings and complete seasonal holdouts, and examine the separate
contributions of calendar, weather, and historical-consumption
covariates. An operational pilot could connect forecast errors to
service-scheduling decisions and measure the resulting energy and cost
outcomes.

\section{Data Availability and
Reproducibility}\label{data-availability-and-reproducibility}

The electricity-consumption and institutional scheduling data were
supplied for the THUAS building study. The data and analysis code will
be publicly released through 4TU. The research archive retains data
checksums, ordered covariates, validation selections, hourly prediction
ledgers, software and hardware records, analysis scripts, and generated
tables and figures. A GitHub repository will accompany publication.

\section{Supplementary Materials}\label{supplementary-materials}

\subsection{Covariate Catalogue}\label{sec-covariate-catalogue}

\begin{longtable}[]{@{}
  >{\raggedright\arraybackslash}p{(\linewidth - 6\tabcolsep) * \real{0.2500}}
  >{\raggedright\arraybackslash}p{(\linewidth - 6\tabcolsep) * \real{0.2500}}
  >{\raggedright\arraybackslash}p{(\linewidth - 6\tabcolsep) * \real{0.2500}}
  >{\raggedright\arraybackslash}p{(\linewidth - 6\tabcolsep) * \real{0.2500}}@{}}
\caption{Stored names, scientific meanings, and units for the 30
retained covariates. The manuscript uses the scientific names; the
stored names enable
reproduction.}\label{tbl-covariate-catalogue}\tabularnewline
\toprule\noalign{}
\begin{minipage}[b]{\linewidth}\raggedright
Source identifier(s)
\end{minipage} & \begin{minipage}[b]{\linewidth}\raggedright
Scientific meaning
\end{minipage} & \begin{minipage}[b]{\linewidth}\raggedright
Domain
\end{minipage} & \begin{minipage}[b]{\linewidth}\raggedright
Unit or encoding
\end{minipage} \\
\midrule\noalign{}
\endfirsthead
\toprule\noalign{}
\begin{minipage}[b]{\linewidth}\raggedright
Source identifier(s)
\end{minipage} & \begin{minipage}[b]{\linewidth}\raggedright
Scientific meaning
\end{minipage} & \begin{minipage}[b]{\linewidth}\raggedright
Domain
\end{minipage} & \begin{minipage}[b]{\linewidth}\raggedright
Unit or encoding
\end{minipage} \\
\midrule\noalign{}
\endhead
\bottomrule\noalign{}
\endlastfoot
Exam\_weeks & Exam-week indicator & calendar/schedule & binary \\
Opening\_hours & Opening-hours indicator & calendar/schedule & binary \\
Hour & Hour label & calendar/schedule & hour \\
Year & Calendar year & calendar/schedule & year \\
Month & Calendar month & calendar/schedule & month \\
Week & ISO week number & calendar/schedule & week \\
Day & Day of month & calendar/schedule & day \\
DayofWeek & Day of week & calendar/schedule & day \\
Month\_sin, Month\_cos & Cyclic month encoding & calendar/schedule &
unitless \\
Week\_sin, Week\_cos & Cyclic week encoding & calendar/schedule &
unitless \\
Day\_sin, Day\_cos & Cyclic day encoding & calendar/schedule &
unitless \\
DayOfWeek\_sin, DayOfWeek\_cos & Cyclic weekday encoding &
calendar/schedule & unitless \\
Hour\_sin, Hour\_cos & Cyclic hour encoding & calendar/schedule &
unitless \\
Work\_week & Work-week indicator & calendar/schedule & binary \\
Weekend & Weekend indicator & calendar/schedule & binary \\
Holiday & Institutional holiday/closure indicator & calendar/schedule &
binary \\
Temperature & Air temperature & weather & °C \\
Temperature\_dew & Dew-point temperature & weather & °C \\
Relative\_humidity & Relative humidity & weather & \% \\
Total\_precipitation & Total precipitation & weather & mm per hourly
interval \\
Pressure & Sea-level pressure & weather & hPa \\
Wind\_speed & Wind speed & weather & m s$^{-1}$ \\
Wind\_direction & Wind direction & weather & degrees \\
lag\_24\_hours & Consumption lag, 24 h & observed consumption & kWh \\
lag\_168\_hours & Consumption lag, 168 h & observed consumption & kWh \\
\end{longtable}

\subsection{Validation and Selected
Settings}\label{sec-validation-settings}

\begin{longtable}[]{@{}
  >{\raggedright\arraybackslash}p{(\linewidth - 10\tabcolsep) * \real{0.14}}
  >{\raggedright\arraybackslash}p{(\linewidth - 10\tabcolsep) * \real{0.12}}
  >{\raggedright\arraybackslash}p{(\linewidth - 10\tabcolsep) * \real{0.13}}
  >{\raggedright\arraybackslash}p{(\linewidth - 10\tabcolsep) * \real{0.15}}
  >{\raggedright\arraybackslash}p{(\linewidth - 10\tabcolsep) * \real{0.12}}
  >{\raggedright\arraybackslash\let\arxivunderscore\_\def\_{\arxivunderscore\allowbreak}}p{(\linewidth - 10\tabcolsep) * \real{0.34}}@{}}
\caption{Validation scores and selected settings for the benchmark
configurations. Every candidate was evaluated on 2,160 hourly validation
targets.}\label{tbl-validation-settings}\tabularnewline
\toprule\noalign{}
\begin{minipage}[b]{\linewidth}\raggedright
Model
\end{minipage} & \begin{minipage}[b]{\linewidth}\raggedright
Validation trials
\end{minipage} & \begin{minipage}[b]{\linewidth}\raggedright
Validation MAE (kWh)
\end{minipage} & \begin{minipage}[b]{\linewidth}\raggedright
Validation RMSE (kWh)
\end{minipage} & \begin{minipage}[b]{\linewidth}\raggedright
Validation R²
\end{minipage} & \begin{minipage}[b]{\linewidth}\raggedright
Selected parameter details
\end{minipage} \\
\midrule\noalign{}
\endfirsthead
\toprule\noalign{}
\begin{minipage}[b]{\linewidth}\raggedright
Model
\end{minipage} & \begin{minipage}[b]{\linewidth}\raggedright
Validation trials
\end{minipage} & \begin{minipage}[b]{\linewidth}\raggedright
Validation MAE (kWh)
\end{minipage} & \begin{minipage}[b]{\linewidth}\raggedright
Validation RMSE (kWh)
\end{minipage} & \begin{minipage}[b]{\linewidth}\raggedright
Validation R²
\end{minipage} & \begin{minipage}[b]{\linewidth}\raggedright
Selected parameter details
\end{minipage} \\
\midrule\noalign{}
\endhead
\bottomrule\noalign{}
\endlastfoot
TimesFM 3.0 with covariates & 64 & 19.98 & 28.90 & 0.861 &
context\_values=8760; make\_positive=False;
point\_forecast=median\_quantile\_index\_4; return\_quantiles=True;
use\_symmetric\_averaging=True; use\_znorm=False \\
TimesFM 3.0 target-only & 64 & 19.37 & 28.64 & 0.863 &
context\_values=8760; make\_positive=False;
point\_forecast=median\_quantile\_index\_4; return\_quantiles=True;
use\_symmetric\_averaging=True; use\_znorm=False \\
TimesFM 2.5 target-only & 8 & 23.38 & 36.26 & 0.781 &
context\_values=672; normalize\_inputs=False \\
XGBoost & 96 & 21.27 & 28.76 & 0.862 & colsample\_bytree=0.6239;
learning\_rate=0.01401; max\_depth=7; min\_child\_weight=0.999;
n\_estimators=1350; reg\_alpha=0.4845; reg\_lambda=0.05573;
subsample=0.7332 \\
Prophet with regressors & 96 & 24.13 & 32.16 & 0.827 &
changepoint\_prior\_scale=0.01784; changepoint\_range=0.8846;
daily\_fourier\_order=10; daily\_mode=additive;
holidays\_prior\_scale=4.843; regressor\_prior\_scale=0.02489;
seasonality\_prior\_scale=0.1609; weekly\_fourier\_order=7;
weekly\_mode=additive \\
Ridge & 25 & 26.82 & 35.69 & 0.787 & alpha=215.4 \\
LSTM + ridge residual & 48 & 30.32 & 42.75 & 0.695 &
residual\_feature\_scaling=standard; residual\_history\_rows=1440;
ridge\_alpha=655.8 \\
Prophet base & 96 & 35.40 & 45.61 & 0.653 &
changepoint\_prior\_scale=0.0144; changepoint\_range=0.8424;
daily\_fourier\_order=12; daily\_mode=additive;
holidays\_prior\_scale=0.854; seasonality\_prior\_scale=4.112;
weekly\_fourier\_order=7; weekly\_mode=additive \\
LSTM base & 72 & 32.08 & 45.91 & 0.648 &
architecture\_profile=\{`dropout': 0.05, `l2': 1e-06, `layers': 1,
`units': 64\}; lookback\_hours=168; optimizer\_profile=\{`batch\_size':
32, `learning\_rate': 0.0006, `optimizer': `rmsprop'\} \\
Seasonal-naive 168 h & 1 & 38.03 & 58.46 & 0.430 & lag\_hours=168 \\
Seasonal-naive 24 h & 1 & 40.19 & 63.29 & 0.332 & lag\_hours=24 \\
\end{longtable}

\subsection{Complete Day-Type
Comparison}\label{sec-full-daytype-comparison}

\begin{figure}

\centering{

\includegraphics[width=1\linewidth,height=\textheight,keepaspectratio,alt={Six aligned panels provide the complete MAE and RMSE comparison for all 11 configurations across workdays, weekends, and holidays.}]{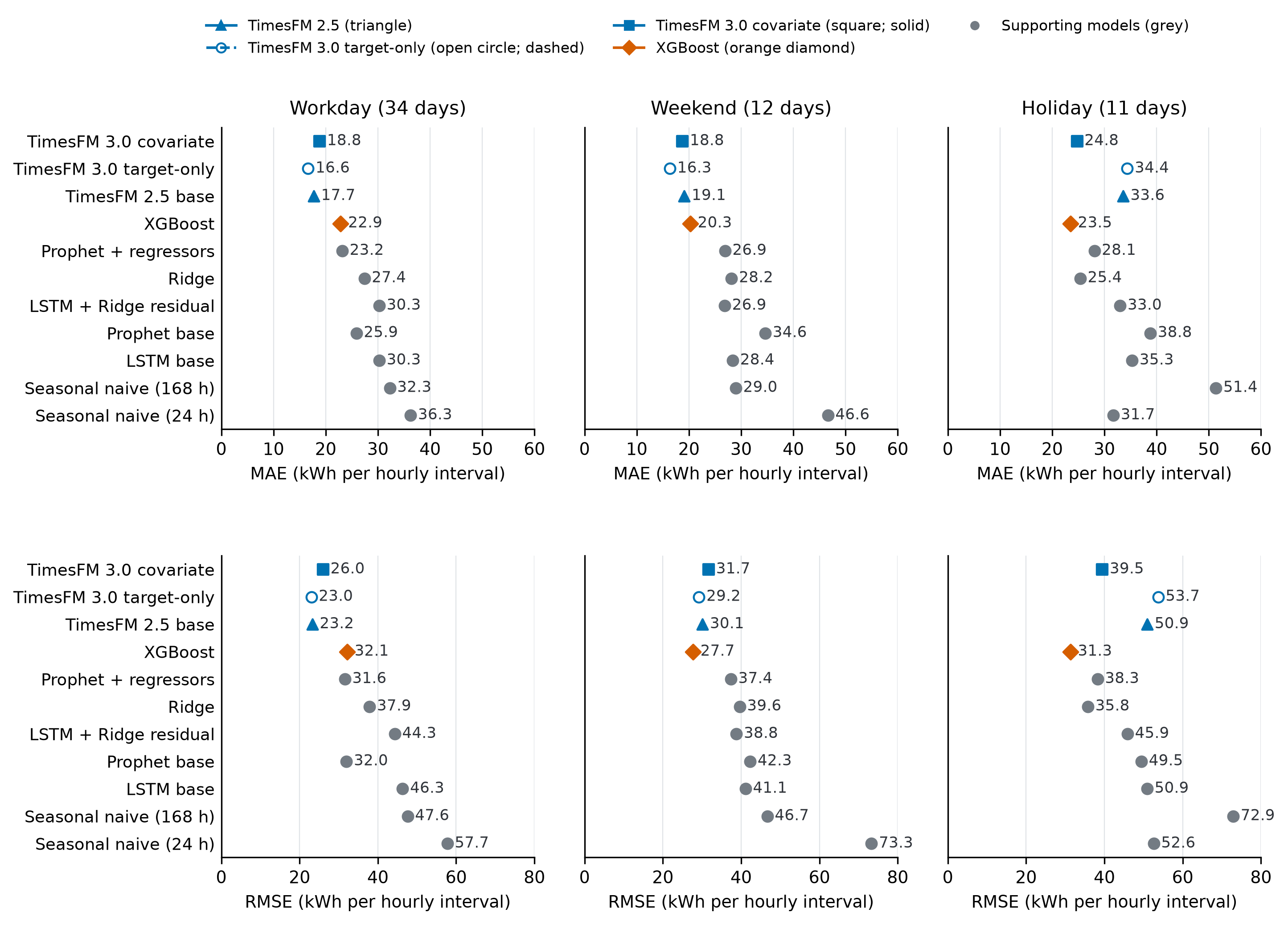}

}

\caption{\label{fig-all-daytype-errors}MAE and RMSE for all 11
configurations on workdays, weekends, and holidays. Model ordering is
identical across panels, and each metric uses the same scale across day
types. Counts are 34 workdays, 12 weekend days, and 11 holidays.}

\end{figure}

Figure~\ref{fig-all-daytype-errors} extends the focused comparison in
Figure~\ref{fig-regime-errors} to all 11 configurations. The workday,
weekend, and holiday groups use the same forecast windows and fixed
model settings as the main benchmark.

\subsection{Additional TimesFM 2.5 Hybrid
Experiments}\label{sec-timesfm25-hybrids}

We compare two jointly tuned TimesFM 2.5 hybrids with the target-only
base. In the first, ridge predicts residuals of TimesFM forecasts; in
the second, TimesFM forecasts residuals of ridge predictions. Context
and residual-regression settings are selected together using the 90
validation origins. The joint-search design was finalised after initial
benchmark results had been inspected, so these additional comparisons
are exploratory.

\begin{longtable}[]{@{}
  >{\raggedright\arraybackslash}p{(\linewidth - 8\tabcolsep) * \real{0.2000}}
  >{\raggedright\arraybackslash}p{(\linewidth - 8\tabcolsep) * \real{0.2000}}
  >{\raggedright\arraybackslash}p{(\linewidth - 8\tabcolsep) * \real{0.2000}}
  >{\raggedright\arraybackslash}p{(\linewidth - 8\tabcolsep) * \real{0.2000}}
  >{\raggedright\arraybackslash}p{(\linewidth - 8\tabcolsep) * \real{0.2000}}@{}}
\caption{Validation and test errors for TimesFM 2.5 target-only and the
two jointly tuned hybrids. Validation covers 2,160 target hours and
testing 1,368; all errors are in
kWh.}\label{tbl-timesfm25-hybrids}\tabularnewline
\toprule\noalign{}
\begin{minipage}[b]{\linewidth}\raggedright
Configuration
\end{minipage} & \begin{minipage}[b]{\linewidth}\raggedright
Validation MAE (kWh)
\end{minipage} & \begin{minipage}[b]{\linewidth}\raggedright
Validation RMSE (kWh)
\end{minipage} & \begin{minipage}[b]{\linewidth}\raggedright
Test MAE (kWh)
\end{minipage} & \begin{minipage}[b]{\linewidth}\raggedright
Test RMSE (kWh)
\end{minipage} \\
\midrule\noalign{}
\endfirsthead
\toprule\noalign{}
\begin{minipage}[b]{\linewidth}\raggedright
Configuration
\end{minipage} & \begin{minipage}[b]{\linewidth}\raggedright
Validation MAE (kWh)
\end{minipage} & \begin{minipage}[b]{\linewidth}\raggedright
Validation RMSE (kWh)
\end{minipage} & \begin{minipage}[b]{\linewidth}\raggedright
Test MAE (kWh)
\end{minipage} & \begin{minipage}[b]{\linewidth}\raggedright
Test RMSE (kWh)
\end{minipage} \\
\midrule\noalign{}
\endhead
\bottomrule\noalign{}
\endlastfoot
TimesFM 2.5 target-only & 23.38 & 36.26 & 21.07 & 31.85 \\
TimesFM 2.5 + ridge residual (jointly tuned) & 23.77 & 36.30 & 20.88 &
31.44 \\
Ridge + TimesFM 2.5 residual (jointly tuned) & 22.54 & 31.66 & 25.14 &
34.91 \\
\end{longtable}

Adding ridge correction to TimesFM gives a small test improvement over
the base: MAE decreases from 21.07 to 20.88 kWh and RMSE from 31.85 to
31.44 kWh. Reversing the order produces the lowest validation RMSE of
the three configurations but higher test errors, with MAE 25.14 kWh and
RMSE 34.91 kWh. Residual modelling therefore provides no consistent
improvement across the two hybrid constructions.

\subsection{Time Convention and Weather
Alignment}\label{sec-time-weather-alignment}

We convert the supplied fixed UTC+1 timestamp labels to UTC by
subtracting one hour, preserving all rows and 24-hour forecast windows.
Calendar covariates and workday/holiday membership retain the
institutional operating-clock labels. Thus, a forecast beginning at
23:00 UTC covers portions of two UTC dates, while monthly and seasonal
error groups follow the individual UTC target timestamps.

For the supplementary weather comparison, a target labelled \(u\) is
assigned to the interval \([u,u+1\,\mathrm{h})\). Temperature, humidity,
pressure, and wind are represented at the interval endpoint;
precipitation is accumulated over that interval. The KNMI workbook key
marks the start of the observation interval, so ERA5 and IFS use the
next valid-time row relative to that key. This mapping was checked
across all 57 context and forecast windows. Station wind averages over a
short observation period, while the gridded wind field is instantaneous.

The original benchmark and extended historical evaluation preserve the
inherited source join. Under the adopted fixed UTC+1 and interval-start
interpretation, that join selects KNMI observations one hourly row later
than the supplementary endpoint mapping. This is a difference between
timestamp assignments, not evidence of incorrect KNMI measurements.
Independent documentation of whether the original meter timestamp
denotes an interval start or end was unavailable. We therefore keep the
two result sets distinct and state their conventions explicitly.

Wind direction is a numerical covariate in degrees, not a binary weather
category. The KNMI value 990 denotes variable direction rather than an
angle; preprocessing treats it as missing and substitutes the median
from observations preceding the forecast origin. The separate
weather-code covariate is excluded from all 30-covariate configurations.
In the supplementary comparison, pressure is mean-sea-level pressure for
all sources.

Open-Meteo documents the historical IFS forecast archive (Open-Meteo,
n.d.-a). Our downloaded records retain weather valid times but not
complete model-initialisation and publication times for each value. This
is sufficient to identify the supplied forecast product, but not to
reconstruct exactly which run an operator could have accessed at every
historical forecast origin.

\subsection{Historical Evaluation and Weather
Breakdowns}\label{sec-supplementary-breakdowns}

The historical tables report target-hour groups. Workday, weekend, and
holiday membership follows the source operating day. Calendar months and
meteorological seasons follow UTC target timestamps, so a forecast
window may contribute hours to two adjacent groups. Seasonal coverage is
unequal; in particular, summer is observed only in 2023.

\begin{longtable}[]{@{}
  >{\raggedright\arraybackslash}p{(\linewidth - 10\tabcolsep) * \real{0.1667}}
  >{\raggedright\arraybackslash}p{(\linewidth - 10\tabcolsep) * \real{0.1667}}
  >{\raggedright\arraybackslash}p{(\linewidth - 10\tabcolsep) * \real{0.1667}}
  >{\raggedright\arraybackslash}p{(\linewidth - 10\tabcolsep) * \real{0.1667}}
  >{\raggedright\arraybackslash}p{(\linewidth - 10\tabcolsep) * \real{0.1667}}
  >{\raggedright\arraybackslash}p{(\linewidth - 10\tabcolsep) * \real{0.1667}}@{}}
\caption{Historical evaluation by day type and meteorological season.
MAE and RMSE are in kWh; target-hour counts identify the coverage of
each group.}\label{tbl-replay-groups}\tabularnewline
\toprule\noalign{}
\begin{minipage}[b]{\linewidth}\raggedright
Group
\end{minipage} & \begin{minipage}[b]{\linewidth}\raggedright
Level
\end{minipage} & \begin{minipage}[b]{\linewidth}\raggedright
Model
\end{minipage} & \begin{minipage}[b]{\linewidth}\raggedright
Target hours
\end{minipage} & \begin{minipage}[b]{\linewidth}\raggedright
MAE (kWh)
\end{minipage} & \begin{minipage}[b]{\linewidth}\raggedright
RMSE (kWh)
\end{minipage} \\
\midrule\noalign{}
\endfirsthead
\toprule\noalign{}
\begin{minipage}[b]{\linewidth}\raggedright
Group
\end{minipage} & \begin{minipage}[b]{\linewidth}\raggedright
Level
\end{minipage} & \begin{minipage}[b]{\linewidth}\raggedright
Model
\end{minipage} & \begin{minipage}[b]{\linewidth}\raggedright
Target hours
\end{minipage} & \begin{minipage}[b]{\linewidth}\raggedright
MAE (kWh)
\end{minipage} & \begin{minipage}[b]{\linewidth}\raggedright
RMSE (kWh)
\end{minipage} \\
\midrule\noalign{}
\endhead
\bottomrule\noalign{}
\endlastfoot
Day type & holiday & TimesFM 3.0 with covariates & 2760 & 17.887 &
34.969 \\
Day type & weekend & TimesFM 3.0 with covariates & 2904 & 14.648 &
25.227 \\
Day type & workday & TimesFM 3.0 with covariates & 7920 & 16.863 &
23.81 \\
Day type & holiday & Seasonal-naive 168 h & 2760 & 40.62 & 69.059 \\
Day type & weekend & Seasonal-naive 168 h & 2904 & 20.574 & 35.139 \\
Day type & workday & Seasonal-naive 168 h & 7920 & 28.317 & 44.806 \\
UTC season & Winter (DJF) & TimesFM 3.0 with covariates & 4344 & 20.247
& 29.47 \\
UTC season & Summer (JJA) & TimesFM 3.0 with covariates & 2208 & 9.893 &
14.243 \\
UTC season & Spring (MAM) & TimesFM 3.0 with covariates & 2927 & 19.915
& 33.975 \\
UTC season & Autumn (SON) & TimesFM 3.0 with covariates & 4105 & 13.976
& 22.692 \\
UTC season & Winter (DJF) & Seasonal-naive 168 h & 4344 & 41.681 &
63.009 \\
UTC season & Summer (JJA) & Seasonal-naive 168 h & 2208 & 15.006 &
24.204 \\
UTC season & Spring (MAM) & Seasonal-naive 168 h & 2927 & 31.717 &
53.167 \\
UTC season & Autumn (SON) & Seasonal-naive 168 h & 4105 & 21.704 &
37.702 \\
\end{longtable}

\begin{longtable}[]{@{}
  >{\raggedright\arraybackslash}p{(\linewidth - 14\tabcolsep) * \real{0.1250}}
  >{\raggedright\arraybackslash}p{(\linewidth - 14\tabcolsep) * \real{0.1250}}
  >{\raggedright\arraybackslash}p{(\linewidth - 14\tabcolsep) * \real{0.1250}}
  >{\raggedright\arraybackslash}p{(\linewidth - 14\tabcolsep) * \real{0.1250}}
  >{\raggedright\arraybackslash}p{(\linewidth - 14\tabcolsep) * \real{0.1250}}
  >{\raggedright\arraybackslash}p{(\linewidth - 14\tabcolsep) * \real{0.1250}}
  >{\raggedright\arraybackslash}p{(\linewidth - 14\tabcolsep) * \real{0.1250}}
  >{\raggedright\arraybackslash}p{(\linewidth - 14\tabcolsep) * \real{0.1250}}@{}}
\caption{Chronological monthly errors and context coverage. Context
ranges are the minimum and maximum numbers of prior hourly observations
supplied to TimesFM for forecasts contributing to the UTC
month.}\label{tbl-replay-months}\tabularnewline
\toprule\noalign{}
\begin{minipage}[b]{\linewidth}\raggedright
UTC month
\end{minipage} & \begin{minipage}[b]{\linewidth}\raggedright
Target hours
\end{minipage} & \begin{minipage}[b]{\linewidth}\raggedright
Context min
\end{minipage} & \begin{minipage}[b]{\linewidth}\raggedright
Context max
\end{minipage} & \begin{minipage}[b]{\linewidth}\raggedright
MAE (kWh)
\end{minipage} & \begin{minipage}[b]{\linewidth}\raggedright
RMSE (kWh)
\end{minipage} & \begin{minipage}[b]{\linewidth}\raggedright
Naive MAE (kWh)
\end{minipage} & \begin{minipage}[b]{\linewidth}\raggedright
Naive RMSE (kWh)
\end{minipage} \\
\midrule\noalign{}
\endfirsthead
\toprule\noalign{}
\begin{minipage}[b]{\linewidth}\raggedright
UTC month
\end{minipage} & \begin{minipage}[b]{\linewidth}\raggedright
Target hours
\end{minipage} & \begin{minipage}[b]{\linewidth}\raggedright
Context min
\end{minipage} & \begin{minipage}[b]{\linewidth}\raggedright
Context max
\end{minipage} & \begin{minipage}[b]{\linewidth}\raggedright
MAE (kWh)
\end{minipage} & \begin{minipage}[b]{\linewidth}\raggedright
RMSE (kWh)
\end{minipage} & \begin{minipage}[b]{\linewidth}\raggedright
Naive MAE (kWh)
\end{minipage} & \begin{minipage}[b]{\linewidth}\raggedright
Naive RMSE (kWh)
\end{minipage} \\
\midrule\noalign{}
\endhead
\bottomrule\noalign{}
\endlastfoot
2022-09 & 457 & 168 & 624 & 10.519 & 14.674 & 10.82 & 14.428 \\
2022-10 & 744 & 624 & 1368 & 10.169 & 19.31 & 18.163 & 35.337 \\
2022-11 & 720 & 1368 & 2088 & 19.474 & 31.703 & 30.046 & 45.45 \\
2022-12 & 744 & 2088 & 2832 & 17.181 & 23.717 & 44.034 & 68.687 \\
2023-01 & 744 & 2832 & 3576 & 19.842 & 26.994 & 40.608 & 59.109 \\
2023-02 & 672 & 3576 & 4248 & 18.398 & 27.194 & 30.404 & 46.698 \\
2023-03 & 744 & 4248 & 4992 & 17.274 & 23.133 & 36.743 & 54.228 \\
2023-04 & 720 & 4992 & 5712 & 27.103 & 47.829 & 32.074 & 58.871 \\
2023-05 & 744 & 5712 & 6456 & 18.187 & 34.065 & 32.2 & 59.184 \\
2023-06 & 720 & 6456 & 7176 & 11.936 & 16.583 & 12.607 & 24.28 \\
2023-07 & 744 & 7176 & 7920 & 8.779 & 12.166 & 16.732 & 24.899 \\
2023-08 & 744 & 7920 & 8664 & 9.029 & 13.709 & 15.601 & 23.411 \\
2023-09 & 720 & 8664 & 8760 & 12.192 & 18.383 & 12.672 & 20.581 \\
2023-10 & 744 & 8760 & 8760 & 13.081 & 21.26 & 30.099 & 53.956 \\
2023-11 & 720 & 8760 & 8760 & 17.313 & 24.573 & 24.285 & 34.409 \\
2023-12 & 744 & 8760 & 8760 & 20.47 & 28.851 & 38.871 & 59.88 \\
2024-01 & 744 & 8760 & 8760 & 22.593 & 34.369 & 50.282 & 73.69 \\
2024-02 & 696 & 8760 & 8760 & 22.999 & 34.261 & 45.011 & 65.255 \\
2024-03 & 719 & 8760 & 8760 & 17.239 & 25.571 & 25.658 & 37.003 \\
\end{longtable}

\begin{longtable}[]{@{}
  >{\raggedright\arraybackslash}p{(\linewidth - 8\tabcolsep) * \real{0.2000}}
  >{\raggedright\arraybackslash}p{(\linewidth - 8\tabcolsep) * \real{0.2000}}
  >{\raggedright\arraybackslash}p{(\linewidth - 8\tabcolsep) * \real{0.2000}}
  >{\raggedright\arraybackslash}p{(\linewidth - 8\tabcolsep) * \real{0.2000}}
  >{\raggedright\arraybackslash}p{(\linewidth - 8\tabcolsep) * \real{0.2000}}@{}}
\caption{Paired daily-MAE differences between weather sources, in kWh,
with 95\% percentile intervals from 10,000 resamples of the 57
forecast-day blocks. Positive values favour the second source
named.}\label{tbl-weather-paired}\tabularnewline
\toprule\noalign{}
\begin{minipage}[b]{\linewidth}\raggedright
Contrast
\end{minipage} & \begin{minipage}[b]{\linewidth}\raggedright
Origin days
\end{minipage} & \begin{minipage}[b]{\linewidth}\raggedright
Estimate (kWh)
\end{minipage} & \begin{minipage}[b]{\linewidth}\raggedright
95\% CI low (kWh)
\end{minipage} & \begin{minipage}[b]{\linewidth}\raggedright
95\% CI high (kWh)
\end{minipage} \\
\midrule\noalign{}
\endfirsthead
\toprule\noalign{}
\begin{minipage}[b]{\linewidth}\raggedright
Contrast
\end{minipage} & \begin{minipage}[b]{\linewidth}\raggedright
Origin days
\end{minipage} & \begin{minipage}[b]{\linewidth}\raggedright
Estimate (kWh)
\end{minipage} & \begin{minipage}[b]{\linewidth}\raggedright
95\% CI low (kWh)
\end{minipage} & \begin{minipage}[b]{\linewidth}\raggedright
95\% CI high (kWh)
\end{minipage} \\
\midrule\noalign{}
\endhead
\bottomrule\noalign{}
\endlastfoot
ERA5 reanalysis − KNMI station 344 & 57 & -0.0002 & -0.1672 & 0.1706 \\
Historical IFS − KNMI station 344 & 57 & 0.1691 & 0.0071 & 0.3305 \\
Historical IFS − ERA5 reanalysis & 57 & 0.1693 & 0.0027 & 0.3249 \\
\end{longtable}

\begin{longtable}[]{@{}lllll@{}}
\caption{Weather-source errors by operating day type. Each source is
evaluated on 816 workday, 288 weekend, and 264 holiday target
hours.}\label{tbl-weather-daytype}\tabularnewline
\toprule\noalign{}
Source & Day type & Target hours & MAE (kWh) & RMSE (kWh) \\
\midrule\noalign{}
\endfirsthead
\toprule\noalign{}
Source & Day type & Target hours & MAE (kWh) & RMSE (kWh) \\
\midrule\noalign{}
\endhead
\bottomrule\noalign{}
\endlastfoot
KNMI station 344 & holiday & 264 & 24.741 & 39.332 \\
KNMI station 344 & weekend & 288 & 18.783 & 31.84 \\
KNMI station 344 & workday & 816 & 18.842 & 26.069 \\
ERA5 reanalysis & holiday & 264 & 24.965 & 39.621 \\
ERA5 reanalysis & weekend & 288 & 18.461 & 31.488 \\
ERA5 reanalysis & workday & 816 & 18.883 & 25.948 \\
Historical IFS & holiday & 264 & 25.291 & 40.118 \\
Historical IFS & weekend & 288 & 18.717 & 31.532 \\
Historical IFS & workday & 816 & 18.971 & 26.051 \\
\end{longtable}

\begin{figure}

\centering{

\includegraphics[width=1\linewidth,height=\textheight,keepaspectratio,alt={Two horizontal dot panels group the three weather sources by day type, showing small source differences alongside larger differences between operating periods.}]{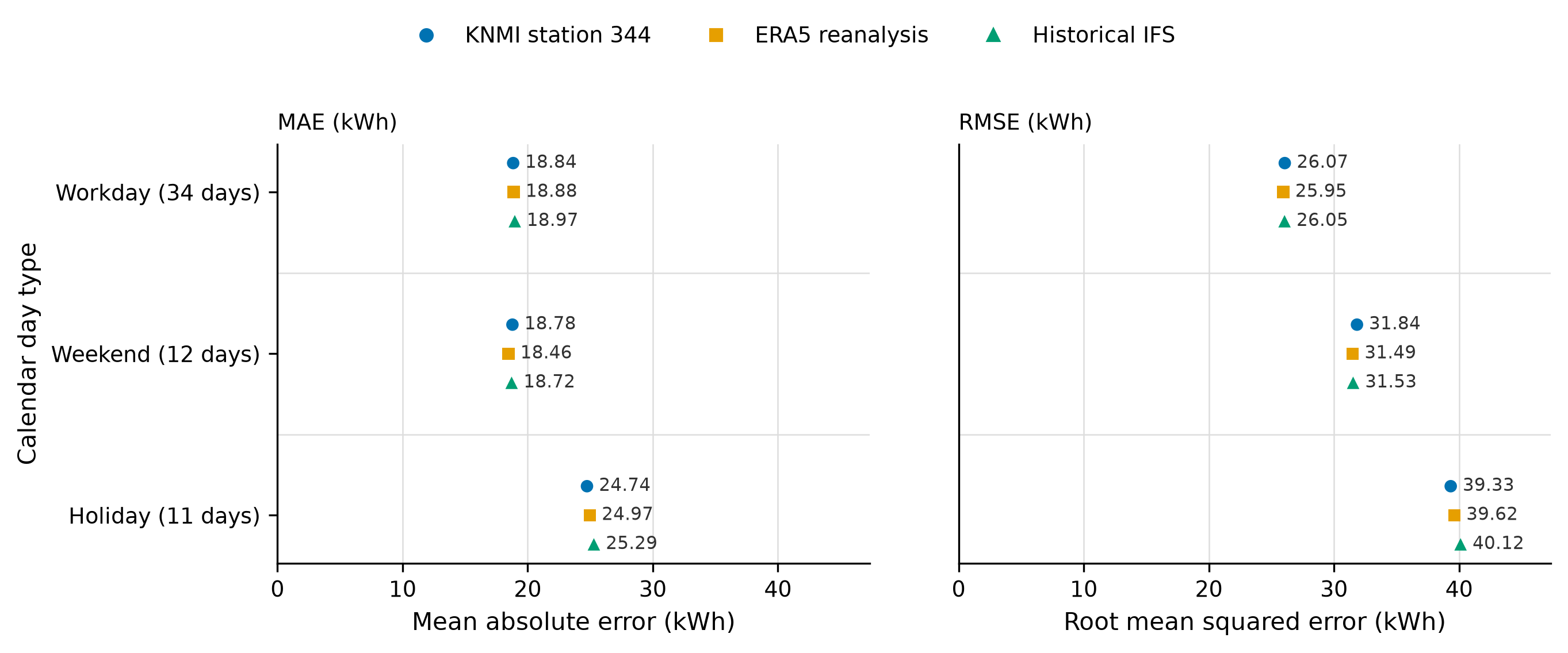}

}

\caption{\label{fig-weather-daytype}MAE and RMSE for each weather source
on workdays, weekends, and holidays. Within each group, distinct markers
identify KNMI, ERA5, and historical IFS on the same error scale. All
sources use the same target hours: 816 workday, 288 weekend, and 264
holiday hours.}

\end{figure}%

\begin{longtable}[]{@{}lllll@{}}
\caption{Weather-source errors by UTC month and meteorological season.
This test contains only February 2024 (649 winter target hours) and
March 2024 (719 spring target hours); month and season therefore
identify the same two
groups.}\label{tbl-weather-month-season}\tabularnewline
\toprule\noalign{}
Source & UTC month / season & Target hours & MAE (kWh) & RMSE (kWh) \\
\midrule\noalign{}
\endfirsthead
\toprule\noalign{}
Source & UTC month / season & Target hours & MAE (kWh) & RMSE (kWh) \\
\midrule\noalign{}
\endhead
\bottomrule\noalign{}
\endlastfoot
KNMI station 344 & February 2024 / DJF & 649 & 23.0108 & 34.7614 \\
KNMI station 344 & March 2024 / MAM & 719 & 17.2218 & 25.5866 \\
ERA5 reanalysis & February 2024 / DJF & 649 & 23.061 & 34.7231 \\
ERA5 reanalysis & March 2024 / MAM & 719 & 17.1761 & 25.4828 \\
Historical IFS & February 2024 / DJF & 649 & 23.1571 & 34.7774 \\
Historical IFS & March 2024 / MAM & 719 & 17.4115 & 25.8402 \\
\end{longtable}

\subsection{Supporting Descriptive
Diagnostics}\label{sec-descriptive-diagnostics}

\begin{figure}

\centering{

\includegraphics[width=0.9\linewidth,height=\textheight,keepaspectratio,alt={Autocorrelation stems over 14 days, with the daily and weekly lags highlighted and labelled.}]{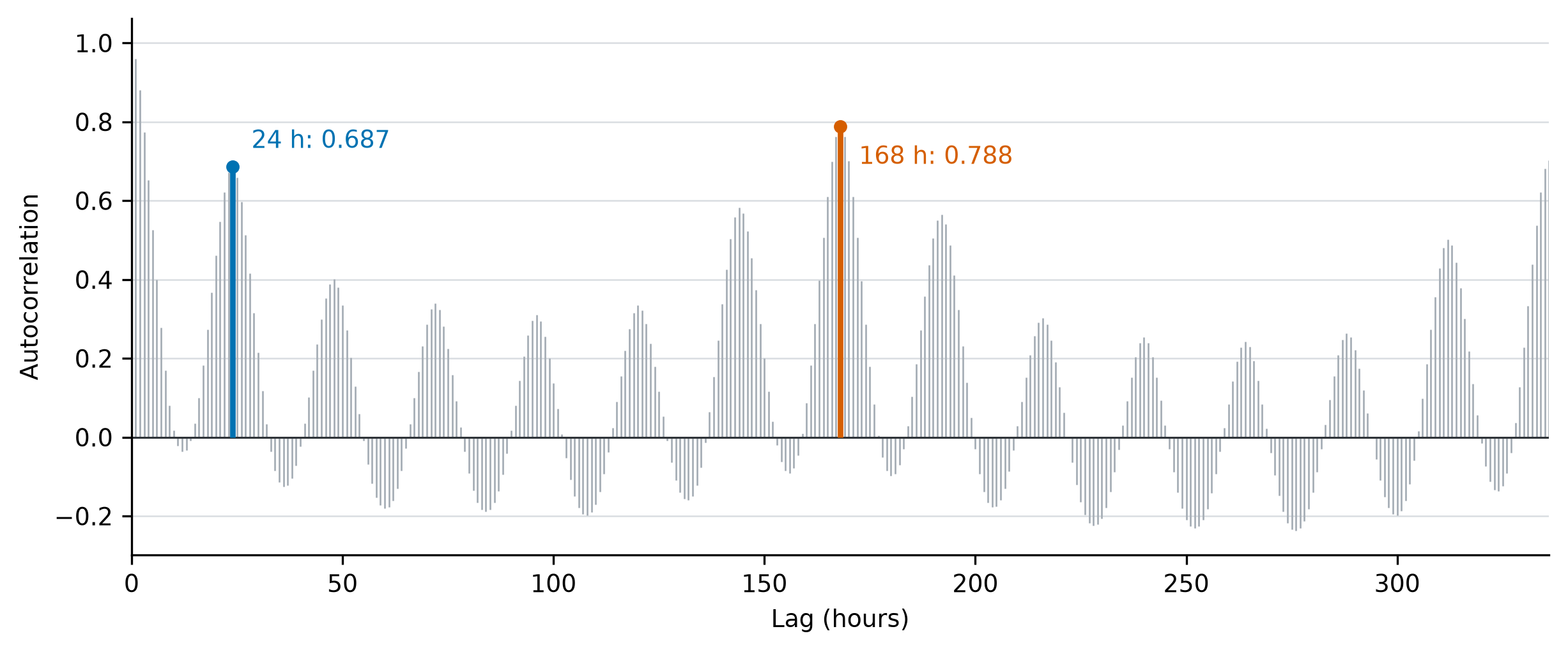}

}

\caption{\label{fig-load-acf}Autocorrelation of hourly electricity
consumption through 336 hours. The highlighted 24- and 168-hour lags
identify daily and weekly recurrence, supporting the seasonal-naive
references.}

\end{figure}

The full-series autocorrelations are 0.687 at 24 hours and 0.788 at 168
hours (Figure~\ref{fig-load-acf}). ADF and KPSS tests give different
conclusions for the raw level because their null hypotheses differ; the
KPSS statistic falls from 1.55 to 0.0046 after first differencing. A
linear trend in daily mean consumption explains little variation
(\(R^2=0.0026\)). We retain these as supporting descriptions rather than
criteria for selecting the final forecast model.

\begin{figure}

\centering{

\includegraphics[width=0.9\linewidth,height=\textheight,keepaspectratio,alt={Mean electricity consumption at each hour over the full observation period.}]{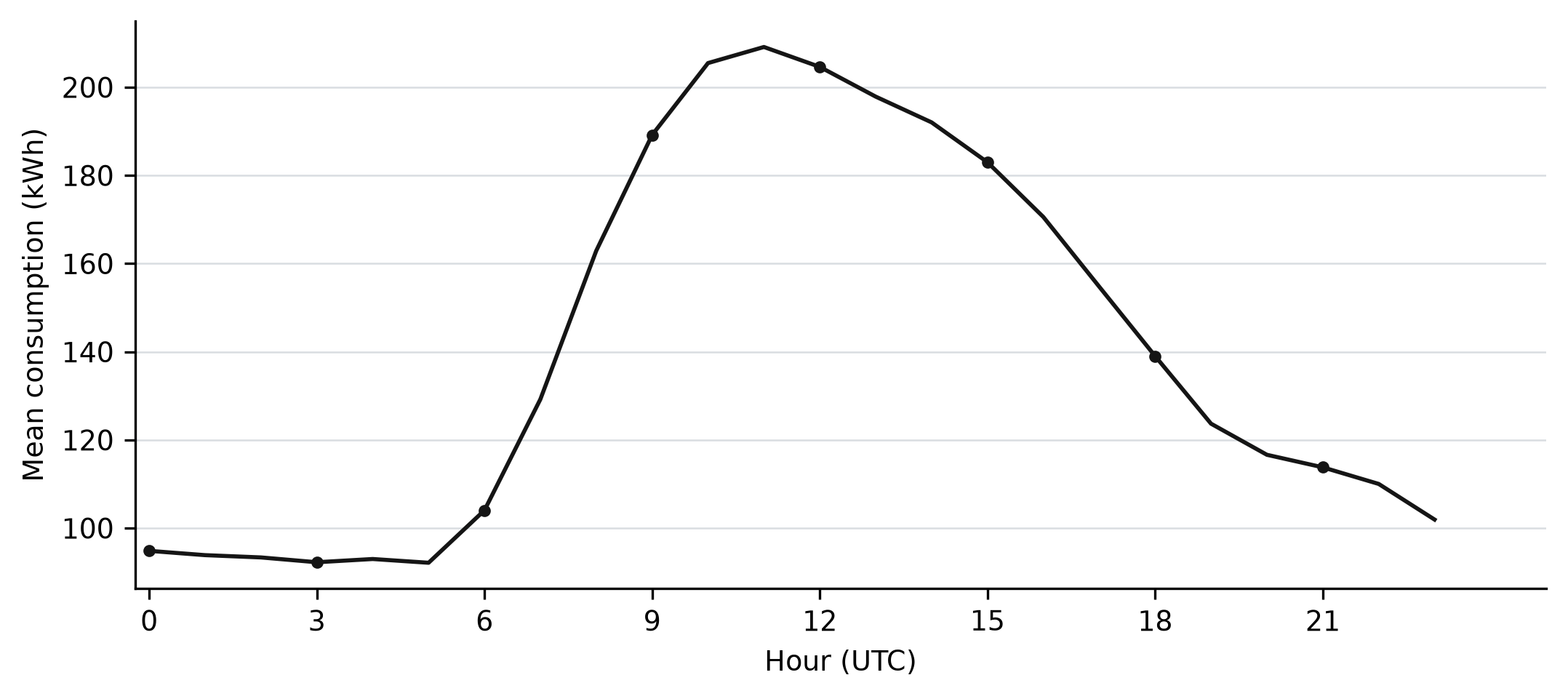}

}

\caption{\label{fig-hourly-profile}Mean hourly electricity consumption
over the complete dataset.}

\end{figure}%

\section*{References}\label{references}
\addcontentsline{toc}{section}{References}

\protect\phantomsection\label{refs}

\end{document}